\def\paperID{2618}
\def\confName{CVPR}
\def\confYear{2024}

\newif\ifreview 
\newif\ifarxiv \newcommand{\arxiv}{\arxivtrue}
\newif\ifcamera 
\newif\ifrebuttal 

\arxiv 

\pdfoutput=1

\documentclass[10pt,twocolumn,letterpaper]{article}
\usepackage{indentfirst}
\usepackage{float}
\usepackage{bbding}
\usepackage{amssymb}
\usepackage{subcaption}
\usepackage{bm}
\usepackage{tikz}

\usepackage[accsupp]{axessibility} 
\usetikzlibrary{spy}
\ifreview \usepackage[review]{cvpr} \fi
\ifarxiv \usepackage[pagenumbers]{cvpr} \fi
\ifrebuttal \usepackage[rebuttal]{cvpr} \fi
\ifcamera \usepackage{cvpr} \fi

\usepackage{graphicx}	
\usepackage{amsmath}	
\usepackage{amssymb}	
\usepackage{booktabs}
\usepackage{times}
\usepackage{microtype}
\usepackage{epsfig}
\usepackage{caption}
\usepackage{float}
\usepackage{placeins}
\usepackage{color, colortbl}
\usepackage{stfloats}
\usepackage{enumitem}
\usepackage{tabularx}
\usepackage{xstring}
\usepackage{multirow}
\usepackage{xspace}
\usepackage{url}
\usepackage{subcaption}
\usepackage{xcolor}
\usepackage[hang,flushmargin]{footmisc}

\ifcamera \usepackage[accsupp]{axessibility} \fi

% ...

\ifarxiv  \fi

\newcommand{\R}[1]{{%
    \textbf{%
        \ifstrequal{#1}{1}{\textcolor{red}{R#1}}{%
        \ifstrequal{#1}{2}{\textcolor{blue}{R#1}}{%
        \ifstrequal{#1}{3}{\textcolor{magenta}{R#1}}{%
        \ifstrequal{#1}{4}{\textcolor{teal}{R#1}}{%
                           \textcolor{cyan}{R#1}%
        }}}}%
    }%
}}

\usepackage{xr-hyper}

\makeatletter
\newcommand*{\addFileDependency}[1]{
  \typeout{(#1)}
  \@addtofilelist{#1}
  \IfFileExists{#1}{}{\typeout{No file #1.}}
}

\makeatother

\definecolor{cvprblue}{rgb}{0.21,0.49,0.74}
\usepackage[pagebackref,breaklinks,colorlinks,citecolor=cvprblue]{hyperref}
\usepackage[capitalize]{cleveref}
\crefname{section}{Sec.}{Secs.}
\crefname{table}{Table}{Tables}
\crefname{figure}{Fig.}{Figs.}

\begin{document}

\newcommand{\rev}[1]{{{#1}}} 

\newcommand{\ba}{\mathbf{a}}
\newcommand{\bb}{\mathbf{b}}
\newcommand{\bc}{\mathbf{c}}
\newcommand{\bd}{\mathbf{d}}
\newcommand{\be}{\mathbf{e}}
\newcommand{\bff}{\mathbf{f}}
\newcommand{\bg}{\mathbf{g}}
\newcommand{\bh}{\mathbf{h}}
\newcommand{\bi}{\mathbf{i}}
\newcommand{\bj}{\mathbf{j}}
\newcommand{\bk}{\mathbf{k}}
\newcommand{\bl}{\mathbf{l}}
\newcommand{\bn}{\mathbf{n}}
\newcommand{\bo}{\mathbf{o}}
\newcommand{\bp}{\mathbf{p}}
\newcommand{\bq}{\mathbf{q}}
\newcommand{\br}{\mathbf{r}}
\newcommand{\bs}{\mathbf{s}}
\newcommand{\bt}{\mathbf{t}}
\newcommand{\bu}{\mathbf{u}}
\newcommand{\bv}{\mathbf{v}}
\newcommand{\bw}{\mathbf{w}}
\newcommand{\bx}{\mathbf{x}}
\newcommand{\by}{\mathbf{y}}
\newcommand{\bz}{\mathbf{z}}
\newcommand{\bA}{\mathbf{A}}
\newcommand{\bB}{\mathbf{B}}
\newcommand{\bC}{\mathbf{C}}
\newcommand{\bD}{\mathbf{D}}
\newcommand{\bE}{\mathbf{E}}
\newcommand{\bF}{\mathbf{F}}
\newcommand{\bG}{\mathbf{G}}
\newcommand{\bH}{\mathbf{H}}
\newcommand{\bI}{\mathbf{I}}
\newcommand{\bJ}{\mathbf{J}}
\newcommand{\bK}{\mathbf{K}}
\newcommand{\bL}{\mathbf{L}}
\newcommand{\bM}{\mathbf{M}}
\newcommand{\bN}{\mathbf{N}}
\newcommand{\bO}{\mathbf{O}}
\newcommand{\bP}{\mathbf{P}}
\newcommand{\bQ}{\mathbf{Q}}
\newcommand{\bR}{\mathbf{R}}
\newcommand{\bS}{\mathbf{S}}
\newcommand{\bT}{\mathbf{T}}
\newcommand{\bU}{\mathbf{U}}
\newcommand{\bV}{\mathbf{V}}
\newcommand{\bW}{\mathbf{W}}
\newcommand{\bX}{\mathbf{X}}
\newcommand{\bY}{\mathbf{Y}}
\newcommand{\bZ}{\mathbf{Z}}
\newcommand{\balpha}{\mbox{\boldmath$\alpha$}}
\newcommand{\bgamma}{\mbox{\boldmath$\gamma$}}
\newcommand{\bGamma}{\mbox{\boldmath$\Gamma$}}
\newcommand{\bmu}{\mbox{\boldmath$\mu$}}
\newcommand{\bphi}{\mbox{\boldmath$\phi$}}
\newcommand{\bPhi}{\mbox{\boldmath$\Phi$}}
\newcommand{\bSigma}{\mbox{\boldmath$\Sigma$}}
\newcommand{\bsigma}{\mbox{\boldmath$\sigma$}}
\newcommand{\btheta}{\mbox{\boldmath$\theta$}}

\newcommand{\mE}{\mathcal{E}}
\newcommand{\mF}{\mathcal{F}}
\newcommand{\mB}{\mathcal{B}}
\newcommand{\mV}{\mathcal{V}}
\newcommand{\mG}{\mathcal{G}}
\newcommand{\mM}{\mathcal{M}}
\newcommand{\mH}{\mathcal{H}}
\newcommand{\mL}{\mathcal{L}}
\newcommand{\mU}{\mathcal{U}}
\newcommand{\mC}{\mathcal{C}}
\newcommand{\mS}{\mathcal{S}}
\newcommand{\mR}{\mathcal{R}}
\newcommand{\mD}{\mathcal{D}}
\newcommand{\mO}{\mathcal{O}}
\newcommand{\mP}{\mathcal{P}}
\newcommand{\mT}{\mathcal{T}}
\newcommand{\mSl}{\mathcal{S}_l}
\newcommand{\mN}{\mathcal{N}}
\newcommand{\mDll}{\mathcal{D}_{l,l'}}

\newcommand{\ra}{\rightarrow}
\newcommand{\la}{\leftarrow}

\def\A{{\cal A}}
\def\B{{\cal B}}
\def\C{{\cal C}}
\def\D{{\cal D}}
\def\E{{\cal E}}
\def\F{{\cal F}}
\def\G{{\cal G}}
\def\H{{\cal H}}
\def\I{{\cal I}}
\def\J{{\cal J}}
\def\K{{\cal K}}
\def\L{{\cal L}}
\def\M{{\cal M}}
\def\N{{\cal N}}
\def\O{{\cal O}}
\def\P{{\cal P}}
\def\Q{{\cal Q}}
\def\R{{\cal R}}
\def\S{{\cal S}}
\def\T{{\cal T}}
\def\U{{\cal U}}
\def\V{{\cal V}}
\def\W{{\cal W}}
\def\X{{\cal X}}
\def\Y{{\cal Y}}
\def\Z{{\cal Z}}
\def\Re{{\mathbb R}}
\def\Cx{{\mathbb C}}
\def\Ze{{\mathbb Z}}
\def\Na{{\mathbb N}}
\def\ud{\mathrm{d}}
\def\eps{\varepsilon}
\def\dist{\textrm{dist}}


\title{NECA: Neural Customizable Human Avatar}

\author{
    Junjin Xiao$^1$\quad  
    Qing Zhang$^{1,3*}$\quad  
    Zhan Xu$^2$\quad  
    Wei-Shi Zheng$^{1,3}$ \\
    \small$^1$School of Computer Science and Engineering, Sun Yat-sen University, China \quad 
    $^2$Adobe Research\\
    \small$^3$Key Laboratory of Machine Intelligence and Advanced Computing, Ministry of Education, China\\
    \tt\small 
    xiaojj37@mail2.sysu.edu.cn\quad 
    zhangq93@mail.sysu.edu.cn\\
    \tt\small
    zhaxu@adobe.com\quad 
    wszheng@ieee.org
}

\twocolumn[{
\renewcommand\twocolumn[1][]{#1}%
\maketitle
\vspace{-5ex}
\begin{center}
\includegraphics[width=0.96\textwidth]{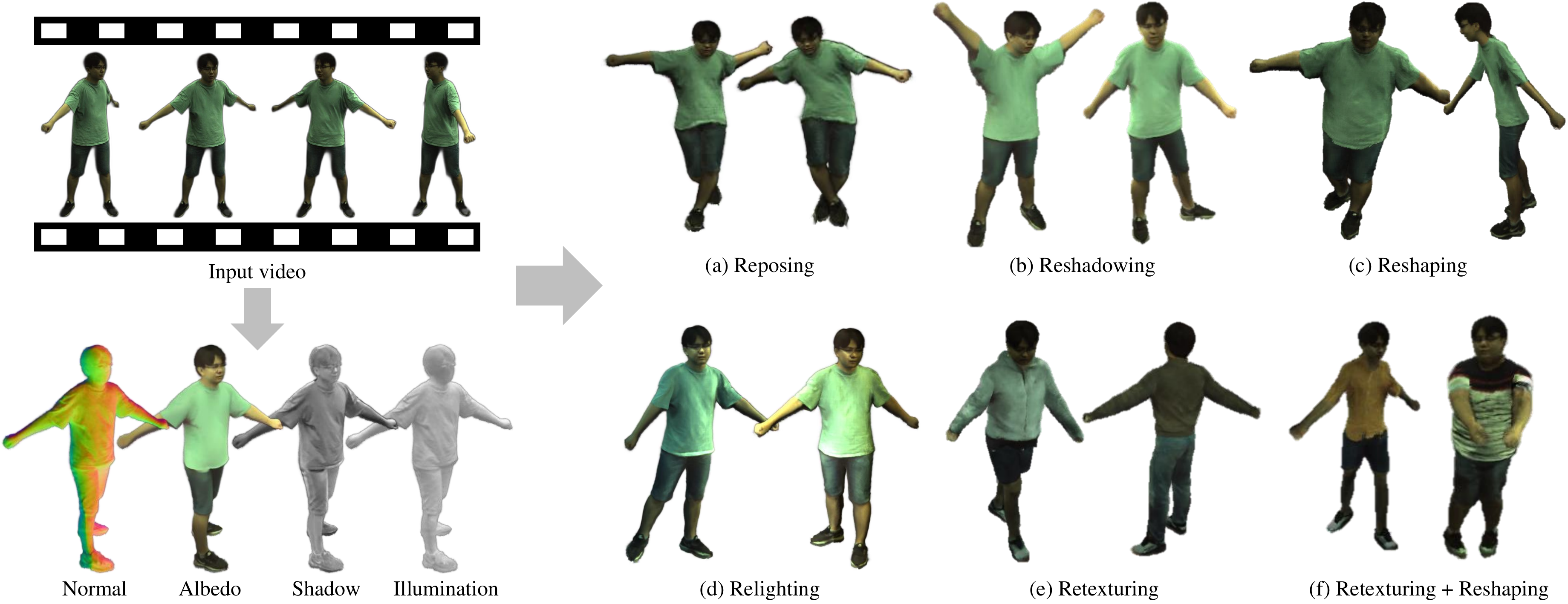}
\captionof{figure}{\textbf{Neural customizable human avatar.} Our method takes as input monocular or sparse multi-view videos and outputs disentangled human representations, including normal, albedo, shadow, and illumination. Such disentanglement enables control of the learned human avatar with arbitrary poses/viewpoints and various customization options such as adjusting shape, shadow, lighting and texture.}
\end{center}
}]

 \maketitle

\setlength{\parindent}{1pc}

\renewcommand{\thefootnote}{}
\footnotetext{$^*$Corresponding author.}

\begin{abstract}

Human avatar has become a novel type of 3D asset with various applications. Ideally, a human avatar should be fully customizable to accommodate different settings and environments. In this work, we introduce NECA, an approach capable of learning versatile human representation from monocular or sparse-view videos, enabling granular customization across aspects such as pose, shadow, shape, lighting and texture. The core of our approach is to represent humans in complementary dual spaces and predict disentangled neural fields of geometry, albedo, shadow, as well as an external lighting, from which we are able to derive realistic rendering with high-frequency details via volumetric rendering. Extensive experiments demonstrate the advantage of our method over the state-of-the-art methods in photorealistic rendering, as well as various editing tasks such as novel pose synthesis and relighting. The code is available at \url{https://github.com/iSEE-Laboratory/NECA}.

\end{abstract}

\vspace{-4mm}
\section{Introduction}
\label{sec:intro}

There is widespread demand for human avatars in many emerging applications such as the metaverse, telepresence, and 3D games. Among them, a basic common requirement for human avatars is that they should be fully editable to allow easy customization across aspects like pose, shape, lighting, texture, and even shadow. While there are numerous works on neural human avatar modelling, as illustrated in~\cref{table:istd}, they are mostly tailored for either animation or relighting purposes, failing to offer full customization capabilities for avatars thus being limited in overall practicality. 

In this work, we present NECA, a novel framework that allows learning fully customizable neural human avatars with photorealistic rendering under any novel pose, viewpoint and lighting, as well as the ability to edit shape, texture, and shadow. To this end, we propose to learn human representation in both Canonical space and the surface space consisting of UV features and local tangent coordinates, so as to capture high-frequency dynamic variations and the shared characteristics across poses with geometry priors for harvesting high-fidelity novel pose synthesis. Besides, to achieve full customization capabilities, we employ distinct MLPs to predict geometry, albedo and shadow separately, and optimize a spherical environmental lighting. The entire framework is trained in a self-supervised manner, constrained only by photometric losses and normal regularization. 

In summary, the main contributions of this work are:

\begin{itemize}[leftmargin=2em]
 \setlength\itemsep{0.5em}
    \item We present NECA, which to our knowledge, is the first framework that is able to learn fully customizable neural human avatars allowing photorealistic rendering.

    \item We propose to learn human representation in dual spaces for capturing high-frequency motion details and geometry-aware characteristics, and represent avatars as disentangled neural fields with distinct geometry and appearance attributes for flexible control.

    \item Extensive experiments demonstrate the broad editing capabilities of our approach, and our significant improvements over prior state-of-the-arts in various tasks such as novel pose synthesis and relighting.
\end{itemize}

\begin{table}
	\centering
	\caption{\textbf{Comparison of customizable attributes enabled by existing neural human avatar reconstruction methods.} } 
	\vspace{-2mm} 
 \resizebox{\linewidth}{!}{
	\begin{tabular}{lccccc}
		\toprule[1pt]
		\multicolumn{1}{l}{Method} &\multicolumn{1}{c}{Pose} & \multicolumn{1}{c}{Lighting} &\multicolumn{1}{c}{Shadow} & \multicolumn{1}{c}{Shape} & \multicolumn{1}{c}{Texture}\\\midrule
		Neural Body~\cite{peng2021neural} &\XSolidBrush  &\XSolidBrush & \XSolidBrush&\XSolidBrush&\XSolidBrush\\
        Neural Actor~\cite{liu2021neural}&\Checkmark  &\XSolidBrush &\XSolidBrush &\Checkmark&\Checkmark\\
		Ani-NeRF~\cite{peng2021animatable}& \Checkmark &\XSolidBrush & \XSolidBrush&\XSolidBrush&\XSolidBrush\\
        DS-NeRF~\cite{zhi2022dual}&\Checkmark  &\XSolidBrush &\XSolidBrush &\Checkmark&\Checkmark\\
        SA-NeRF~\cite{Xu_2022_CVPR}&\Checkmark  &\XSolidBrush &\XSolidBrush &\Checkmark&\Checkmark\\
        HumanNeRF~\cite{weng_humannerf_2022_cvpr}&\Checkmark  &\XSolidBrush &\XSolidBrush &\XSolidBrush&\XSolidBrush\\
        ARAH~\cite{ARAH:2022:ECCV}&\Checkmark  &\XSolidBrush & \XSolidBrush&\XSolidBrush&\XSolidBrush\\
        TAVA~\cite{li2022tava}&\Checkmark  &\XSolidBrush & \XSolidBrush&\XSolidBrush&\Checkmark\\
        Relighting4D~\cite{chen2022relighting}& \XSolidBrush &\Checkmark & \XSolidBrush&\XSolidBrush&\XSolidBrush\\
        MonoHuman~\cite{yu2023monohuman}&\Checkmark  &\XSolidBrush & \XSolidBrush&\XSolidBrush&\XSolidBrush\\
        CustomHumans~\cite{ho2023custom}&\Checkmark  &\XSolidBrush & \XSolidBrush&\Checkmark&\Checkmark\\
        UV Volumes~\cite{Chen_2023_CVPR}&\Checkmark  &\XSolidBrush & \XSolidBrush&\Checkmark&\Checkmark\\
        PoseVocab~\cite{li2023posevocab}&\Checkmark  &\XSolidBrush & \XSolidBrush&\XSolidBrush&\XSolidBrush\\
        Sun~\etal~\cite{Sun_2023_ICCV}&\Checkmark  &\Checkmark & \XSolidBrush&\XSolidBrush&\XSolidBrush\\
        \midrule
		Ours   &\Checkmark  & \Checkmark& \Checkmark&\Checkmark&\Checkmark\\
        \bottomrule[1pt]
	\end{tabular}
 }
	\vspace{-2mm}
	\label{table:istd}
\end{table}
\section{Related Work}
\label{sec:related}

\noindent \textbf{Human Avatar Reconstruction.} Early methods explicitly represent human avatars with deformable templates such as SMPL \cite{SMPL:2015} and its derivatives \cite{SMPLX:2019,STAR:2020}. Such template-based representation has a fixed mesh topology, and always fails to capture high-fidelity details. In recent years, the advent of neural implicit fields has witnessed the great success in 3D reconstruction \cite{Chen_2019_CVPR,wang2021neus,Or-El_2022_CVPR,Park_2019_CVPR,Mescheder_2019_CVPR,mildenhall2020nerf,Sun,Tang}. 
Among them, SDF-based methods \cite{pifuSHNMKL19,he2020geopifu,saito2020pifuhd,zheng2020pamir,xiu2022icon,xiu2023econ} demonstrate promising results in creating lifelike human models. However, they are unable to represent animatable humans. Various later methods \cite{peng2021animatable,liu2021neural,He_2021_ICCV,Huang_2020_CVPR,zhi2022dual} are designed to address this limitation by deforming the human body from observation space to canonical space \cite{weng_humannerf_2022_cvpr,yu2023monohuman}. Particularly, Animatable NeRF \cite{peng2021animatable} proposes to learn per-pose related latent codes for skinning weights in the observation space for deformation, but it requires extra fine-tuning in the inference stage. ARAH \cite{ARAH:2022:ECCV} uses a root-finding algorithm \cite{chen2021snarf} for transformations, thereby enabling novel pose rendering, but at the cost of low rendering speed. Note, as lighting and texture information are usually integrated into a single MLP in these methods, they do not support any appearance editing. There also exist some works that allow appearance editing \cite{Chen_2023_CVPR,ho2023custom,li2022tava,Xu_2022_CVPR}, but they mainly focus on shape, pose, and texture adjustments.   

\vspace{0.5em}
\noindent \textbf{Human Relighting.} A widely adopted technique for relighting human is inverse rendering \cite{Bi_2020_CVPR,Schmitt_2020_CVPR,Yu_2019_CVPR,Pandey_2021_totalRelight,Meka_2020_relightables,Guo_2019_relightables,Kanamori_2018_relightingHuman}, which aims to disentangle geometry, material, and lighting from observed images. However, they are either restricted to novel view and novel pose synthesis because of lacking underlying 3D representation \cite{Pandey_2021_totalRelight,Kanamori_2018_relightingHuman}, or depend heavily on the hard-to-obtain one-light-at-a-time (OLAT) images \cite{Guo_2019_relightables}. Recent implicit-field based methods \cite{Zhang_2021_CVPR,Zhang21nerffactor,Srinivasan_2021_CVPR,Jin2023TensoIR,wang2023fegr,Alldieck_2022_CVPR,Corona_2023_CVPR,Zheng_2023_CVPR} address these limitations by directly learning the underlying 3D structure and unknown illumination from the input images, relying on photometric or geometry loss. While these methods demonstrate promising novel view synthesis results and fine-grained geometry reconstruction, they are mostly tailored for static scenes and not applicable to dynamic humans with complex non-rigid motion and shadow. By extending NeRF \cite{mildenhall2020nerf}, some concurrent works \cite{zhen2023relightable,Sun_2023_ICCV,chen2022relighting} gain applicability to dynamic humans. Nevertheless, they struggle to render high-fidelity humans under arbitrary novel poses \cite{chen2022relighting}, and none of them provide capabilities for texture and shadow editing.

\vspace{-1mm}
\section{Method}
\label{sec:method}
\begin{figure*}[th]
    \centering
    \includegraphics[width=\linewidth]{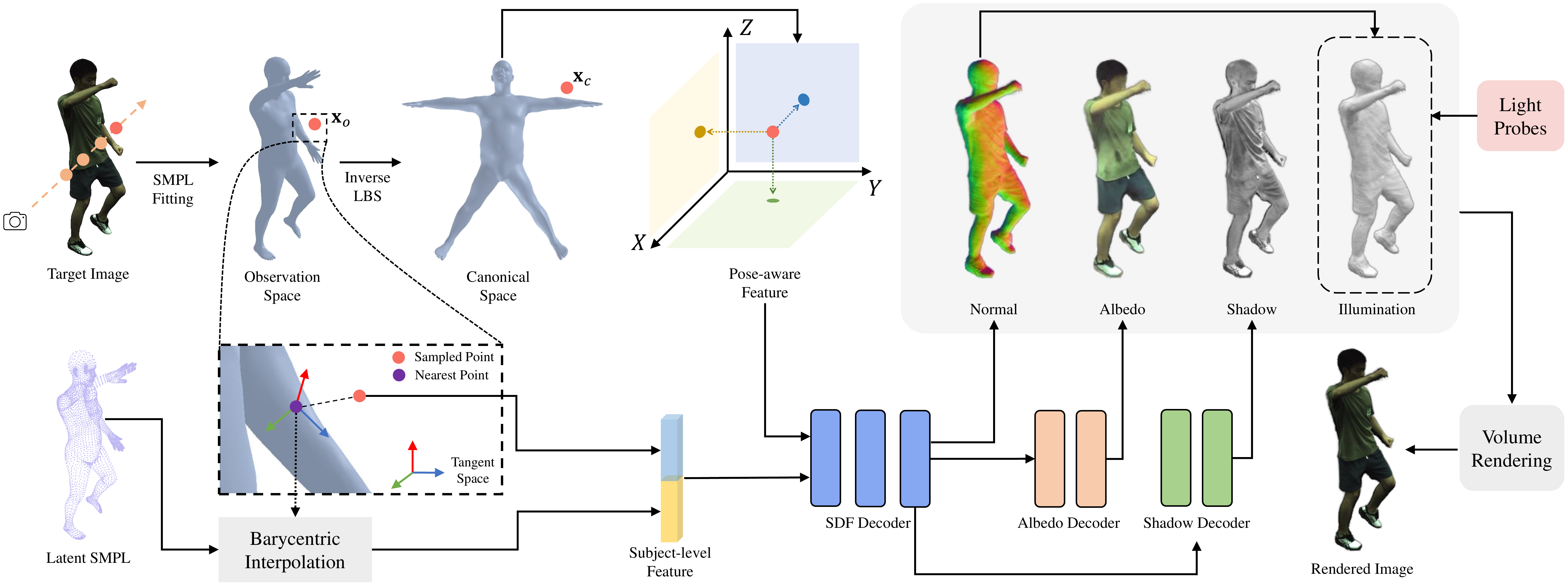}
    \caption{\textbf{Overview of NECA.} We first sample points along the camera ray and transform the query points from observation space to canonical space. Next, we query the pose-aware feature by projecting points to factorized tri-plane that per-pose optimized. Then we construct tangent space of the nearest surface point to the query points on SMPL, and obtain the subject-level feature by concatenating the tangent space local coordinate and the learned latent code in the surface space. Finally, to enable flexible customization, we disentangle the neural fields into attributes including SDF, albedo and shadow, as well as a learnable environmental lighting, by decoding the extracted features with distinct MLPs. The entire network is trained in a self-supervised manner, with only photometric losses and normal regularization.}
    \label{fig:method}
\end{figure*}

Our method aims to generate a fully customizable human avatar from as few as a monocular video. We represent the human avatar in complementary dual spaces (\cref{dual-space}), for capturing both high-frequency pose-aware features and geometry-aware subject-level characteristics. From such representation, we derive the disentangled neural fields, each dedicated to a specific attribute such as SDF, shadow, albedo, as well as environmental lighting (\cref{disentangle}). 
The disentanglement is learned in a self-supervised manner, with only photometric losses and normal regularization (\cref{loss}). Once the avatar has been generated, we can adjust each decomposed attributes separately for diverse applications, ranging from reposing and relighting to reshaping, shadow editing and texture swapping, as demonstrated in our experiments. An overview of our method is shown in \cref{fig:method}.

\subsection{Dual-space Dynamic Human Representation}
\label{dual-space}

\noindent \textbf{Motivation.} Most existing works define neural representation in either Canonical space or the surface space of the deformable human template \cite{SMPL:2015}, each with its own advantages and limitations. Concretely, Canonical space is more generic and offers greater feature capacity to capture high-frequency details, but may lead to disturbing misalignment across poses because of ignoring the shape prior of the human body. In contrast, the surface space leverages the shape prior, focusing on the proximate surface volume. Nevertheless, it relies much on pose prior, which limits the expressiveness of the learned representation. Therefore, we propose to incorporate the strengths of both spaces. We define the pose-aware representation in Canonical space to mitigate the influence of strong pose prior and enhance expressiveness for diverse motion dynamics, and the subject-level representation in the surface space due to its consistency across poses. In a recent work \cite{liu2021neural}, Canonical coordinates and UV features are both utilized in NeRF. They however do not explicitly learn Canonical space feature representation, resulting in relatively limited capability to represent motion variations. Moreover, the surface space feature is only employed for color prediction, leaving the shape prior unexploited. 

\vspace{0.5em}
\noindent \textbf{Canonical space for pose-aware feature.}
Pose changes lead to high-frequency appearance variations, such as dynamics of clothing and shadows resulting from self-occlusion. To better capture such intricate pose-dependent details, We adopt tri-plane representation \cite{shao2023tensor4d,hu2023Tri-MipRF,peng2023representing} to learn pose-aware features. Concretely, we construct tri-plane $\bT_{\btheta}^{\{XY, XZ,YZ\}}\in \mathbb{R}^{L\times  L\times  D}$ for each training pose $\btheta$ in the canonical space.
Given a query point $\bx_o$ in the observation space, we transform it to the canonical space using inverse linear blend skinning \cite{SMPL:2015}:
\begin{equation}
    \bx_c=\sum_b^{N_j} W_b \bB_b^{-1} \bx_o.
\end{equation}
Here, $b$ is index of joints, $N_j$ denotes the total number of joints. $W_b$ denotes the skinning weight of point $\bx_o$ associated with joint $b$, and $\bB_b^{-1}$ is the transformation of joint $b$ from the observation space to the canonical space. We follow the approach outlined in \cite{liu2021neural} to calculate skinning weight $W_b$. We can then fetch the pose-aware feature $\bp_o$ for $\bx_o$ by projecting $\bx_c$ onto three feature planes, and sample corresponding features:
\begin{equation}\label{global_eq} 
\bp_o=\oplus(\bt_{\btheta}^{XY},\bt_{\btheta}^{XZ},\bt_{\btheta}^{YZ},\btheta) ,
\end{equation}
where $\oplus$ denotes concatenation, $\bt_{\btheta}^{\{XY,XZ,YZ\}}$ are the sampled features from the feature planes by bilinear interpolation. 
We append pose $\btheta$ to be more robust.

However, this straightforward implementation incurs significant memory costs due to the per-pose unique tri-plane feature. To reduce memory cost, we employ CP decomposition \cite{Chen2022ECCV,Harshman1970FoundationsOT,Carroll1970AnalysisOI} to factorize the feature planes into orthogonal feature vectors. This is achieved by factorizing the axis-aligned planes into a sum of $R$ outer products of 1D vectors:
\begin{equation} 
\label{tri-plane}
\bT_{\btheta}^{mn}= \sum_{r=1}^R \bv_r^{m} \circ \bv_r^{n} \circ \bv_r^{mn} ,
\end{equation}
where $m,n$ are different axes from $\{X,Y,Z\}$. 
$\bv_r^{\{x,y,z\}}\in\mathbb{R}^{L}$ represent learned axis-aligned vectors,
and $\bv_r^{\{xy, xz, yz\}}\in \mathbb{R}^D$ represent learned vectors in the feature dimension. We dramatically compress the tri-plane, reducing the memory complexity from $\O(L^2)$ to $\O(RL)$ where $R\ll L$.

The learned factorized tri-planes from all training poses form a pose-aware feature dictionary $\{\bT_{\btheta}\}$. During test time, given an unseen pose, we select the top 5 most similar training poses, and average the features obtained from corresponding tri-planes, weighted by the pose similarity. A recent work, PoseVocab \cite{li2023posevocab}, also uses a pose-aware embedding dictionary, but it learns global features for each local joint and its feature representativeness is degraded for achieving better memory efficiency. In our case, pose-aware features are learned with a global pose based on the factorized tri-plane, resulting in features with strong representation power while keeping a low memory footprint.

\begin{figure}[t]  
  \centering            
   \includegraphics[width=0.9\linewidth]{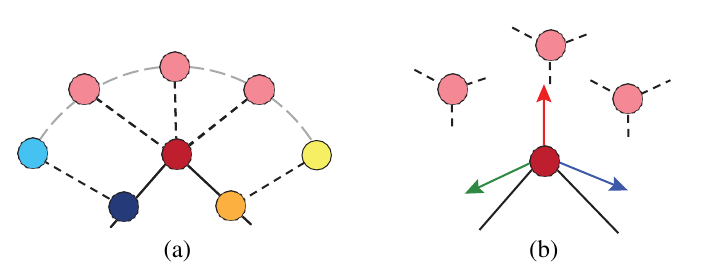}
  \vspace{-6pt}
 \caption{\textbf{Local coordinate used by \cite{zhang2022ndf,ho2023custom,liu2021neural} and the one used by us.} (a) Previous works defined their local coordinate in UV space. The three pink points will be mapped to the same red point on the surface, thus share the same feature. Such many-to-one problem is also mentioned in \cite{Xu_2022_CVPR}. (b) We complement UV space with local tangent space. Although pink points are still mapped to the same surface point, now they have different coordinates in the local tangent space. Their features are therefore different. As previous works, our local coordinate is also rotation invariant.}
 \vspace{-3mm}
\label{fig:tangespace}

\end{figure}

\vspace{0.5em}
 \noindent \textbf{Surface space for subject-level feature.}
\label{surface-space}
Subject-level feature describes pose-agnostic characteristics. We employ the SMPL model \cite{SMPL:2015} as prior to align poses, and define our subject-level feature on its UV map. While earlier methods \cite{zhang2022ndf,ho2023custom,liu2021neural} have adopted the UV volume as pose-agnostic space, their approaches incur the many-to-one problem \cite{Xu_2022_CVPR} illustrated in \cref{fig:tangespace}(a): multiple 3D points are mapped to the same point on edges and vertices, due to the finite area of each triangle. Inspired by \cite{Blinn_1978_Sig}, we address such limitation by constructing tangent space for any point on the SMPL surface, and encode the query 3D point based on its local coordinate in the tangent space of the nearest surface point, as a supplement to the UV feature map.

Specifically, for a surface point $\bx_s$, we construct its associated tangent space by creating the local coordinates from its tangent vector $\bt_s$, bitangent vector $\bb_s$, and normal vector $\bn_s$:
\begin{equation}
    \mM(\bx_s)=[\bt_s,\bb_s,\bn_s]^T .
\end{equation}
More details on calculating TBN matrix $\mathcal{M}_{s}$ can be found in our supplementary material. For each 3D query point $\bx_o$ in the observation space, we first find its nearest surface point $\bx_s^*$ on the SMPL mesh, and then transform the global coordinate of $\bx_o$ to the local tangent space.
\begin{equation}
    \bx_l=\mM(\bx_s^*)(\bx_o - \bx_s^*).
\label{eq:local_tangent_space}
\end{equation}

Eq.~\eqref{eq:local_tangent_space} can differentiate all 3D points given a particular tangent space without the many-to-one problem. The derived $\bx_l$ is invariant to human pose change. To indicate the position of $\bx_s$ and provide geometry awareness, we calculate the associated UV feature $\bg_s$ of the surface point $\bx_s^*$.
\begin{equation}
    \bg_s=\mB_{u_s^*,v_s^*}(\bC_{[\mV_f]}) .
\end{equation}
Here, $\{\bC_v\}$ is a set of learnable latent codes associated with SMPL vertices $v=[1,...,N_v]$. $\mV_f$ are the three vertices of the triangle $\bx_s$ falls in, and $\bC_{[\mV_f]}$ denotes their latent codes. $\mB_{u_s,v_s}(\cdot)$ is the interpolation operation based on the barycentric coordinates $(u_s^*,v_s^*)$.

With both $\bx_l$ and $\bg_s$, our subject-level feature $\bs_o$ of the query point $\bx_o$ is formulated as the concatenation of them. We in practice also encode $\bx_l$ with the positional encoding function introduced in \cite{mildenhall2020nerf}.

\input{figs/compare_on_zju}
\subsection{Disentangled Neural Fields}
\label{disentangle}
The proposed NECA is designed to facilitate diverse customization tasks, including shape, pose, texture, lighting and shadow. Reposing and reshaping are achieved by the adoption of SMPL model. We enable other customization functionalities by decoding the extracted features into distinct neural fields. Unlike \cite{mildenhall2020nerf} which produces density and radiance, our method consumes the canonical position $\bx_c$, the viewing direction $\be$, the pose-aware feature $\bp_o$ and the subject-level feature $\bs_o$ into SDF $d$, albedo $\ba$, shadow $v$, while simultaneously optimizing the representation $\bL$ for lighting. Such an approach expands our editing capability by exposing more controllable attributes for rendering. 

The overall process $(\bx_c, \be, \bs_o, \bp_o) \rightarrow (d, \ba, v)$ is achieved by three separate MLPs. First, the geometry MLP takes all the inputs except for the viewing direction and outputs SDF $d$ and a latent feature $\bh$, similar to \cite{ARAH:2022:ECCV, li2023posevocab}:
\begin{equation}
    d, \bh = \text{MLP}_{geo}(\bx_c, \bs_o, \bp_o; \Theta_{geo}).
\end{equation}

The output SDF can be converted to density $\sigma$ with methods such as \cite{yariv2021volume} for volumetric rendering. Moreover, we can calculate normal vector by normalizing the gradient of SDF $d$ with respect to $\bx_c$ \cite{DE-NeRF, peng2022animatable}:
\begin{equation}\label{normal}
    \bn_c = \cfrac{\partial d}{\partial \bx_c} / ||\cfrac{\partial d}{\partial \bx_c}||_2.
\end{equation}
The normal vector provides critical information about geometry details and orientation, which also benefits our prediction of shadow by another shadow MLP: 
\begin{equation}
    v = \text{MLP}_{shadow}(\be, \bn_c, \bh; \Theta_{shadow}).
\end{equation}
Finally, the albedo MLP predicts the albedo given latent feature $\bh$ and $\bx_c$:
\begin{equation}
    \ba = \text{MLP}_{albedo}(\bx_c, \bh; \Theta_{albedo}).
\end{equation}

Besides the intrinsic characteristics, we also optimize a particular representation of the environmental lighting. We assume the lighting during capturing the input videos is unchanged and in gray-scale, and utilize the learnable light probes \cite{Gao_2022_efficient_lp,chen2022relighting} to represent it in latitude-longitude format as $\bI \in \mathbb{R}^{16 \times 32 \times 1}$. Note that although we assume the light is gray-scale during training, we can replace it with arbitrary light with RGB color during testing time for relighting.

With all the above disentangled neural fields and the lighting, the color of a query point $\bx_o$ can be obtained using the following rendering equation as in \cite{rudnev2022nerfosr,chen2022relighting,rendering_eq}:
\begin{equation}\label{color}
    \bc = \ba \odot v \odot \sum_i I_i (\bn_o \cdot \bm{\omega}_i) \Delta \bm{\omega}_i,
\end{equation}
where $\odot$ denotes the Hadamard Product, $\bn_o$ represents the normal vector transformed to the observation space, $\bm{\omega}_i$ denotes the direction from query points $\bx_o$ in the observation space to the light position, and $(\cdot)$ indicates the dot product.

To render the final color of a pixel, we follow the volumetric rendering \cite{volume_rendering}, by integrating the colors $\{\bc_i,i\in[1, N]\}$ of $N$ sampled query points along the ray $r$ from that pixel position based on their density $\{\sigma_i\}$.
\begin{equation}\label{volume_rendering}
\begin{aligned}
    C(r) &= \sum_{i}^N \exp(-\sum_{j=1}^{i-1} \sigma_j \delta_j) (1-\exp(-\sigma_i \delta_i)) \bc_i,
\end{aligned}
\end{equation}
where $\delta_i$ indicates the distance between two sampled points.

Our disentangled neural fields can produce higher-fidelity rendering while maximizing controllability for customization during rendering. Unlike traditional physically-based rendering approaches which rely on BRDFs for color and explicit visibility for shadows, our method leverages a data-driven approach for inverse rendering. Such choice is more robust to noise, and can better represent complex shadows, such as those caused by clothes wrinkles.

\begin{figure*}[th]
    \centering
    \includegraphics[width=\linewidth]{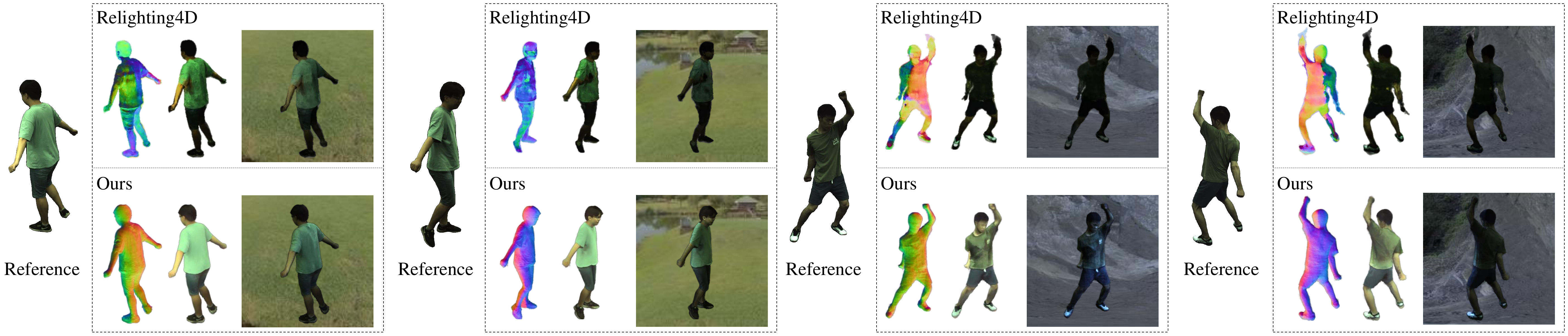}
    \vspace{-18pt}
    \caption{\textbf{Qualitative comparison of relighting under novel pose and view on ZJU-MoCap dataset.} Given the original frame as reference, we compare with Relighting4D on the estimated normal and albedo, as well as the generated relighting results. As shown, our method outperforms Relighting4D in both appearance disentanglement and relighting. Please see the supplementary material for more results.}
    \vspace{-2mm}
\label{fig:com-relight}
\end{figure*}

\subsection{Loss Function}
\label{loss}

We supervise the training of NECA with photometric losses and normal regularization described below. 

The photometric losses measure the consistency of the rendering of NECA to the ground-truth targets. First, we supervise our rendered pixel colors $C(r)$ by minimizing the squared Canonical distance to the target pixel colors $C_{gt}(r)$:
\begin{equation}
    \mathcal{L}_c = \|C_{gt}(r) - C(r)\|_2^2.
\end{equation}

Next, we employ a perceptual loss LPIPS \cite{Zhang_2018_CVPR} with VGG network \cite{vgg} as backbone to enhance the quality of the rendered images:
\begin{equation}
    \mathcal{L}_p = ||VGG_{gt}(r) - VGG(r)||_2^2.
\end{equation}
Note, we apply patch-based ray sampling \cite{weng_humannerf_2022_cvpr} to keep memory efficiency.

The foreground masks serve as supervisory signal for geometry learning. We employ the binary cross-entropy loss to reduce discrepancies in determining whether each ray intersects the subject or not \cite{peng2022animatable,Yariv_2020_NEURIPS}.
\begin{equation}
    \mathcal{L}_m = BCE({\rm sigmoid} (- \rho d_r), M(r)).
\end{equation}
Here, $\rho$ is a scale factor, $d_r$ is the minimal SDF value of ray $r$, and $M(r)$ is the ground truth binary mask. 

We incorporate normal regularization to better capture geometry details, which is also beneficial to the prediction of albedo and shadow. One regularization is the eikonal loss \cite{Gropp_2020_ICML} that encourages the SDF field to be smooth:
\begin{equation}
    \mathcal{L}_e = \left(\left\|\cfrac{\partial d}{\partial \bx_c}\right\|_2 - 1\right)^2.
\end{equation}

We also notice that normals occasionally face backward due to the inherent symmetry of SMPL. To address this, we introduce another normal regularization $\mL_n$:
\begin{equation}
    \mathcal{L}_n = \max(-\bn_c \cdot \bn_s, 0),
\end{equation}
where $\bn_c$ represents the normal vector as defined in Eq.~\eqref{normal}, and $\bn_s$ denotes the normal vector of the nearest SMPL vertex. $\bn_c \cdot \bn_s$ indicates the dot product of the two vectors. 

The overall training objective function for our network is formulated as follows:
\begin{equation}
    \begin{aligned}
    \mathcal{L} =& \mathcal{L}_c + \lambda_p \mathcal{L}_p + \lambda_m \mathcal{L}_m + \lambda_e \mathcal{L}_e +  \lambda_n \mathcal{L}_n,
    \end{aligned}
\end{equation}
where we set $\lambda_p=0.1, \lambda_m=1, \lambda_e=0.1, \lambda_n=0.1$. 

\subsection{Implementation Details}
We implement our network in Pytorch and train it for $200K$ iterations with a mini-batch size of 1 on an NVidia RTX 4080 GPU. The entire network is optimized using the Adam optimizer \cite{adam} with a learning rate decaying exponentially from an initial value of $5 \times 10^{-4}$ to $1 \times 10^{-4}$. For each training batch, we sample 1024 rays and 64 points per ray. The patch size for the perceptual loss is set as 32, with a single patch in use. We set $|\bC_v|\in \mathbb{R}^{16}$ for the latent code attached to SMPL vertex. For tri-plane decomposition, we set the grid resolution $L_x \times L_y \times L_z$ as $512 \times 512 \times 128$, and a lower resolution along z-axis is used for memory efficiency due to the flatness of human body \cite{Chen2023PAMI}. We set $\bv_r^{\{xy, xz, yz\}} \in \mathbb{R}^{32}$, and $R=48$. 

\begin{table}[]
\centering
\resizebox{\linewidth}{!}
{
\begin{tabular}{l  c  c  c  c  c c c }
\toprule[1pt]
\multirow{1}{*}{} & \multicolumn{3}{c}{Novel View} & \multicolumn{3}{c}{Novel Pose} & \multirow{2}{*}{Param.}\\ 
 & PSNR $\uparrow$ & SSIM $\uparrow$ & LPIPS $\downarrow$ & PSNR $\uparrow$ & SSIM $\uparrow$ & LPIPS $\downarrow $ &\\ 
\midrule
NB \cite{peng2021neural}     &28.2&0.944&0.096&23.7&0.893&0.145&4.37M\\
AN \cite{peng2021animatable} &25.8&0.913&0.146&23.2&0.887&0.165&1.96M\\
DS \cite{zhi2022dual}        &27.8&0.937&0.125&23.8&0.892&0.157&0.50M\\
ARAH \cite{ARAH:2022:ECCV}   &28.4&\textbf{0.947}&0.082&24.5&0.909&0.109&87.09M\\
PV \cite{li2023posevocab}    &27.5&0.939&0.068&24.3&0.905&0.095&24.08M\\
Ours                            &\textbf{28.5}&\underline{0.946}&\textbf{0.067}&\textbf{25.0}&\textbf{0.914}&\textbf{0.091}&18.37M\\
\bottomrule[1pt]
\end{tabular}
}
\vspace{-0.5em}
\caption{\textbf{Quantitative comparison of novel pose/view synthesis on all 9 subjects on ZJU-MoCap dataset.} As shown, our method achieves the best results with comparable model size.}

\label{table:zjumocap}
\vspace{-2mm}
\end{table}

\section{Experiments}
\label{sec:experiments}
\begin{figure*}[th]
    \centering
    \includegraphics[width=\linewidth]{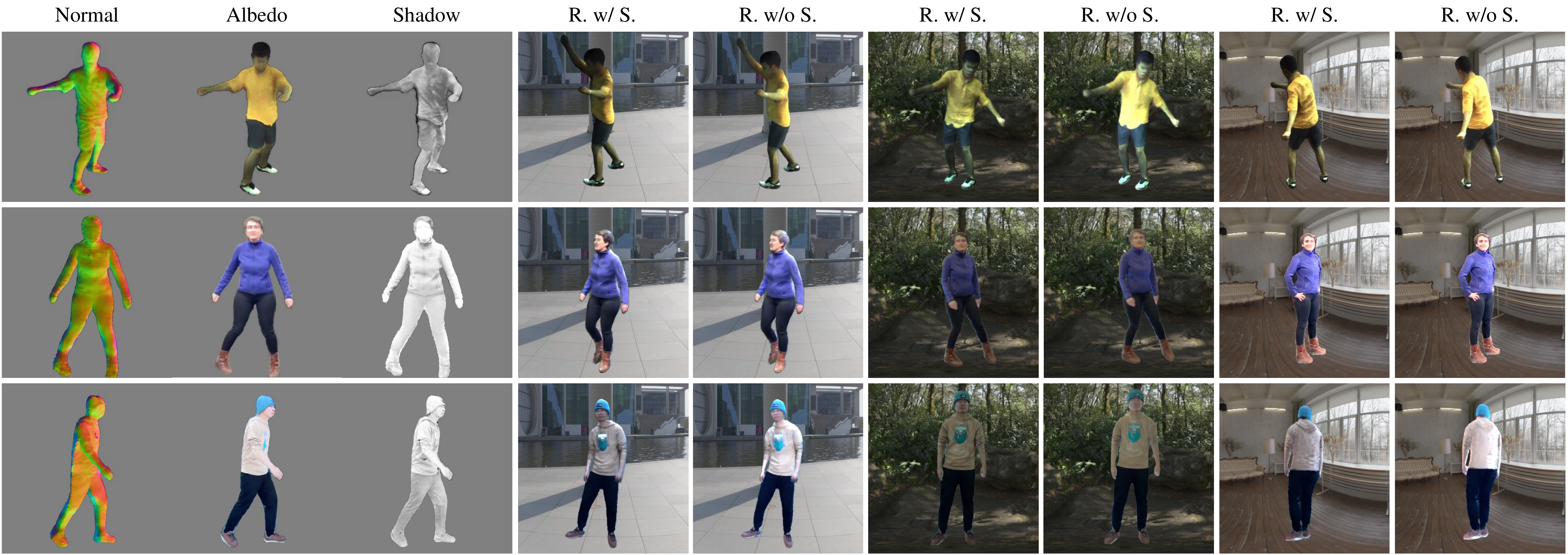}
    \vspace{-18px}
\caption{\textbf{More relighting results on the indoor ZJU-MoCap (top row) and the outdoor NeuMan (bottom two rows) datasets.} ``R. w/ S.'' and ``R. w/o S.'' refer to results with and without shadows, respectively. Our method produces high-fidelity dynamic human renderings that respond faithfully to novel lighting. Please check the supplementary videos for more results.}
\vspace{-2mm}
\label{fig:show-relight}
\end{figure*}

\subsection{Evaluation Datasets and Metrics}

\noindent \textbf{Datasets.} 
We evaluate our method on an indoor dataset ZJU-MoCap \cite{peng2021neural}, an outdoor dataset NeuMan \cite{jiang2022neuman}. ZJU-MoCap consists of 9 sequences with dynamic humans with complex motion captured by multiple cameras in laboratory environment. We train and evaluate our method on all 9 sequences, and follow the training and testing view and pose split introduced in \cite{peng2021neural}. NeuMan consists of videos of approximately 10 to 20 seconds in length, where the ``Bike" and ``Seattle" sequences are employed for training and evaluation. Note, we also evaluate our method on DeepCap~\cite{deepcap} and DynaCap~\cite{habermann2021dynacap} datasets, and quantitatively validate our superiority in relighting using a synthetic dataset. Please see the supplementary material for details.

\begin{figure}[t]  
  \centering    
  \captionsetup[subfigure]{labelformat=empty,labelsep=space}
    \begin{subfigure}[c]{0.0915\textwidth}		\includegraphics[width=1\textwidth, trim=200 200 350 50, clip]{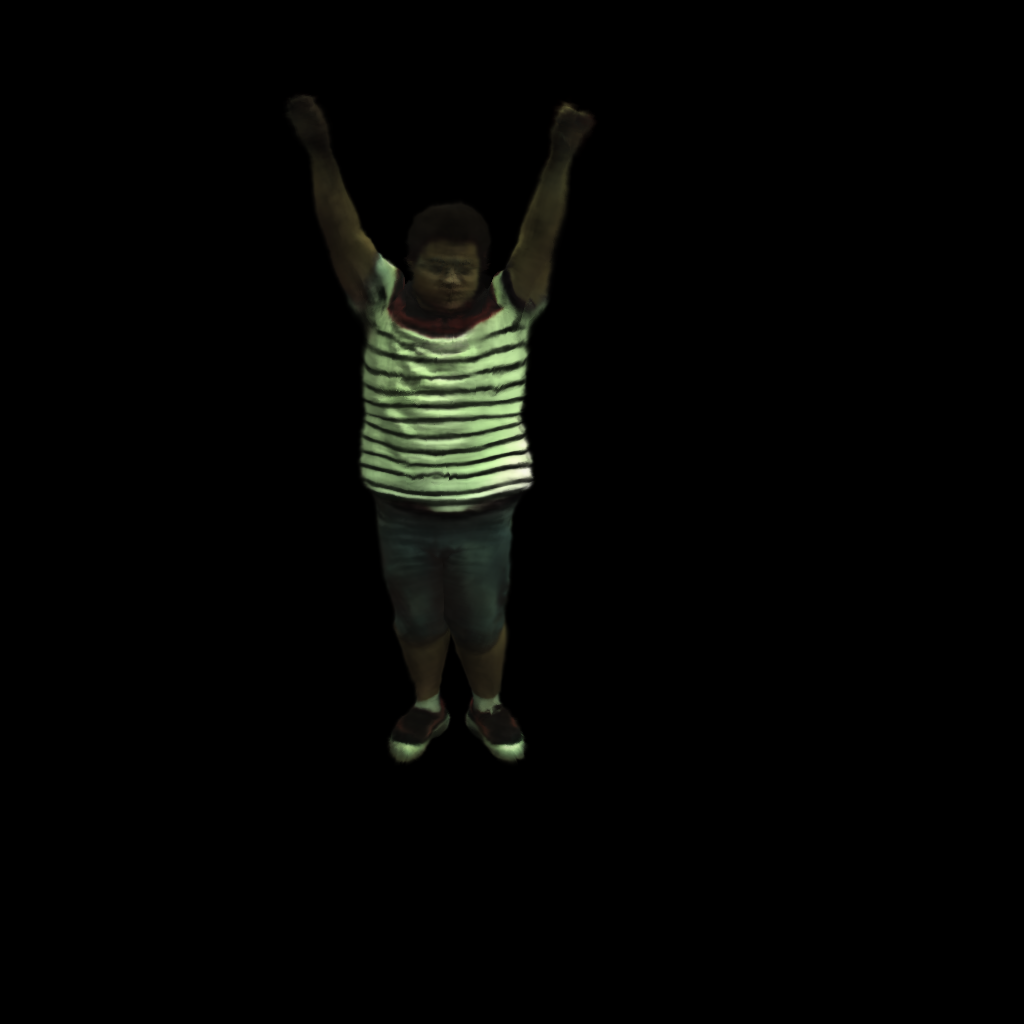}
    \end{subfigure}
    \begin{subfigure}[c]{0.0915\textwidth}
		\includegraphics[width=1\textwidth, trim=200 200 350 50, clip]{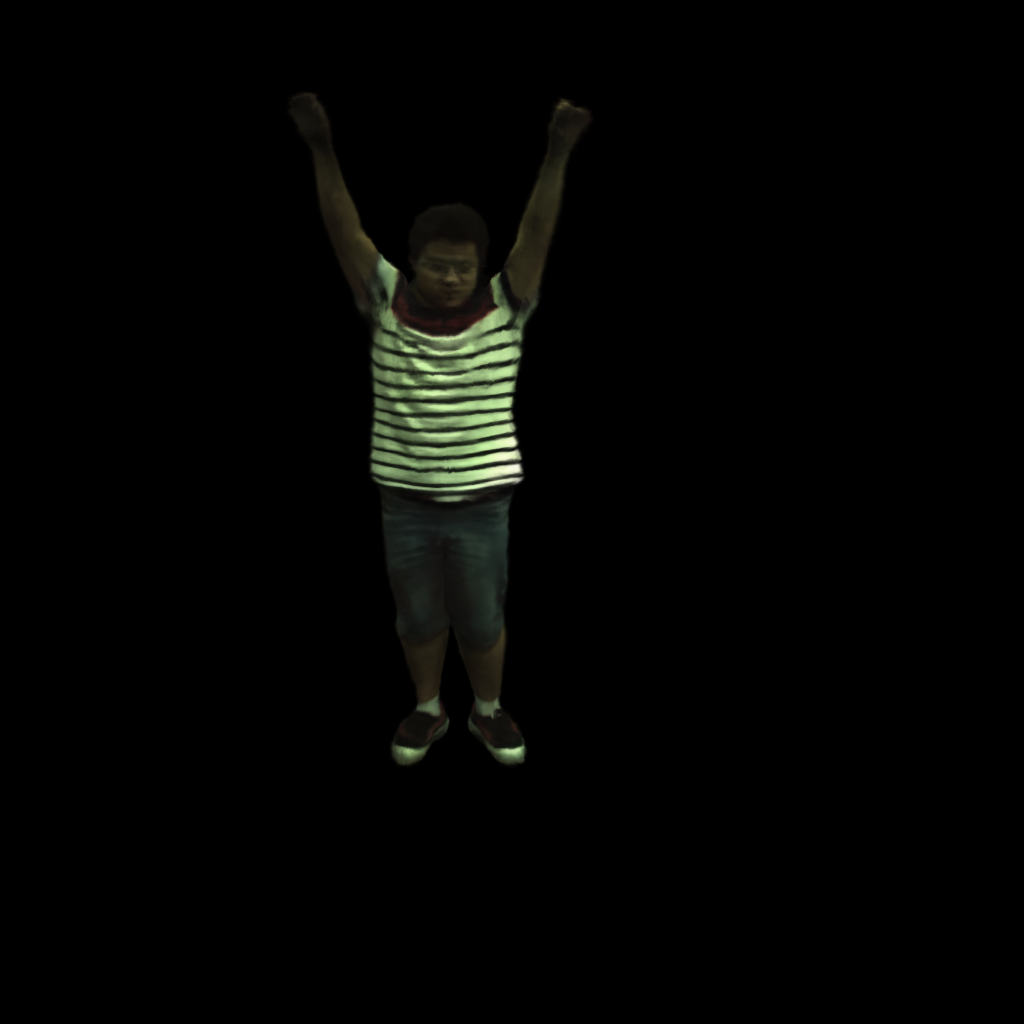}
    \end{subfigure}
    \begin{subfigure}[c]{0.0915\textwidth}
		\includegraphics[width=1\textwidth, trim=200 200 350 50, clip]{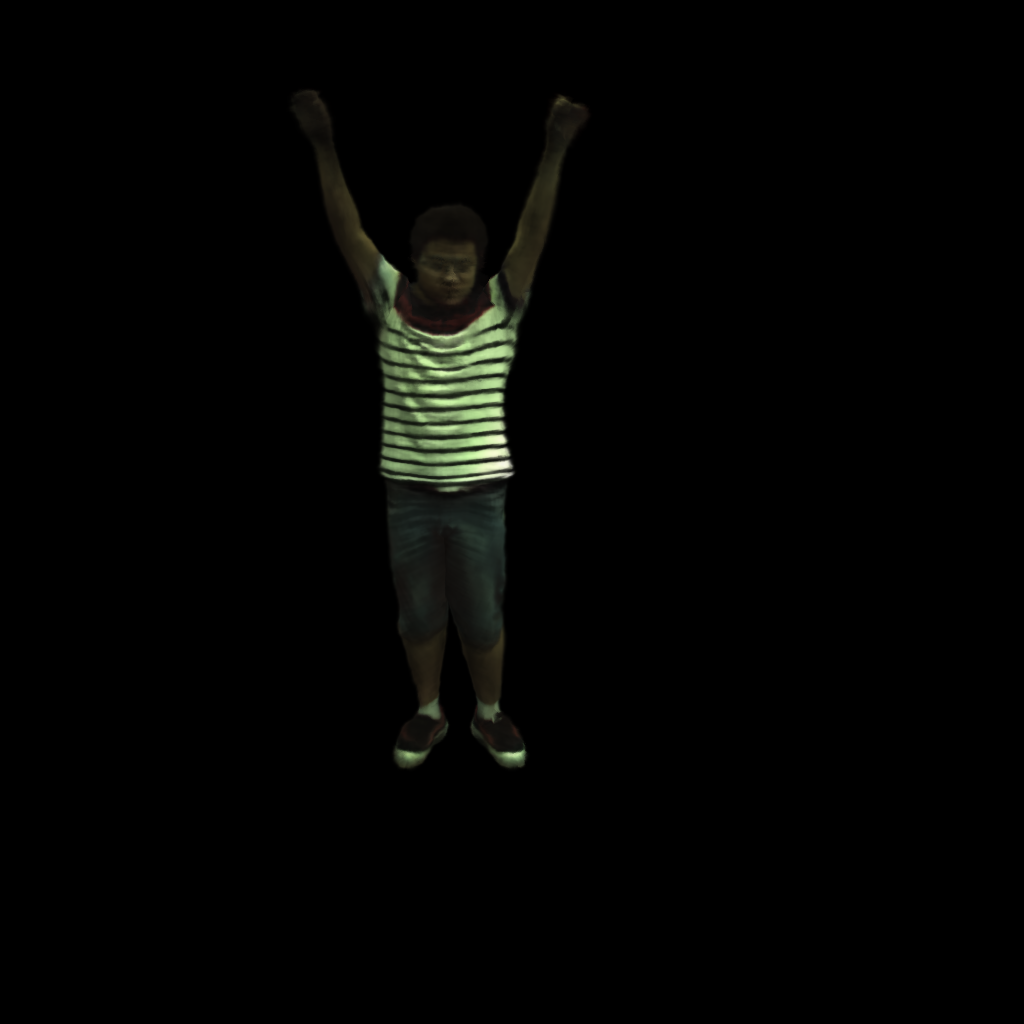}
    \end{subfigure}
    \begin{subfigure}[c]{0.0915\textwidth}
		\includegraphics[width=1\textwidth, trim=200 200 350 50, clip]{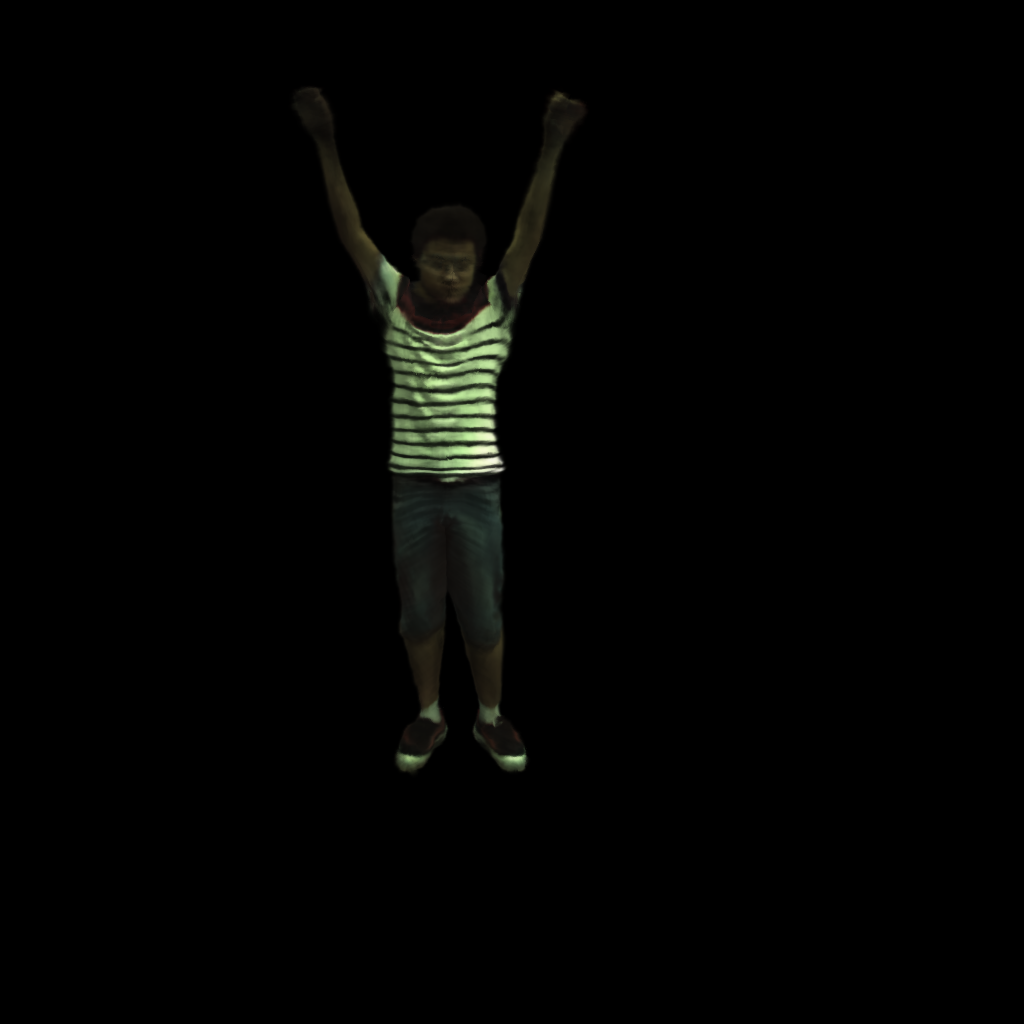}
    \end{subfigure}
    \begin{subfigure}[c]{0.0915\textwidth}
		\includegraphics[width=1\textwidth, trim=200 200 350 50, clip]{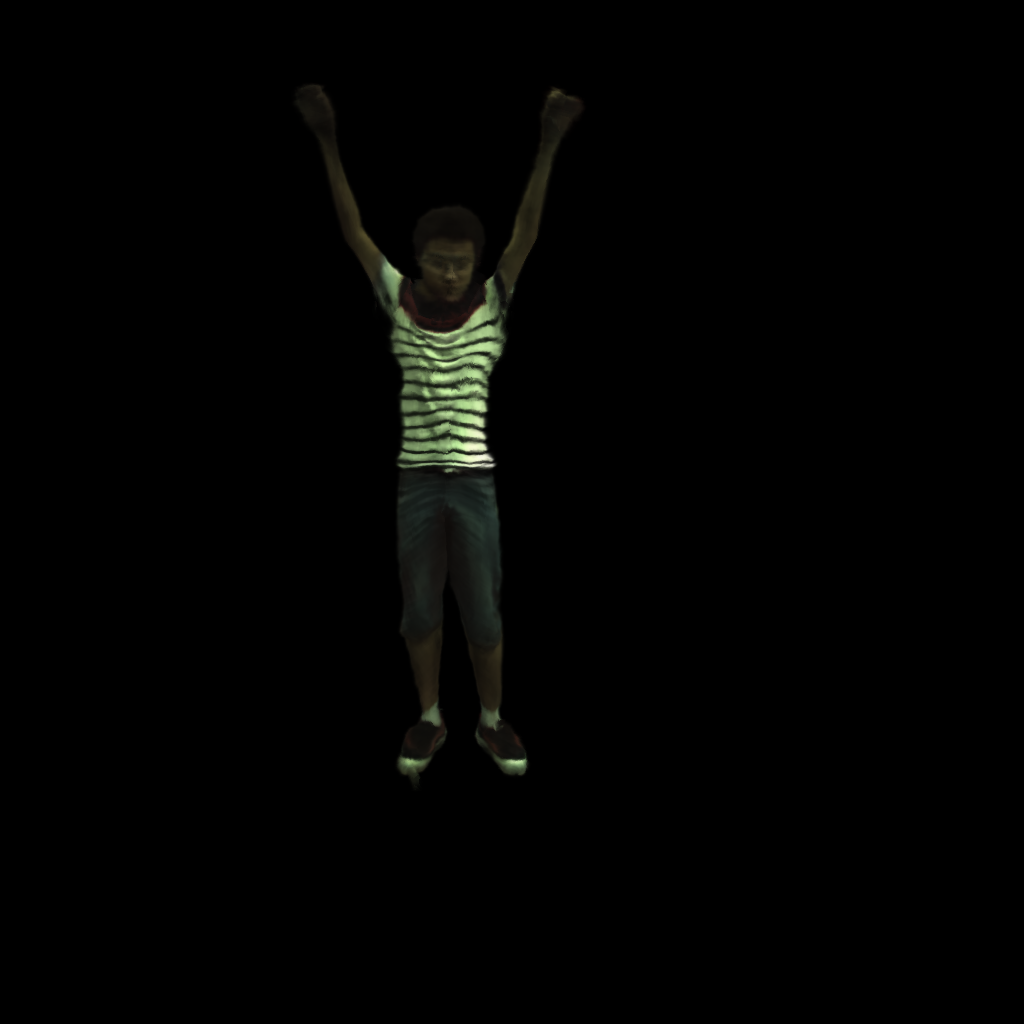}
    \end{subfigure}
    \vspace{1pt} \\
    \begin{subfigure}[c]{0.0915\textwidth}
		\includegraphics[width=1\textwidth, trim=200 200 350 50, clip]{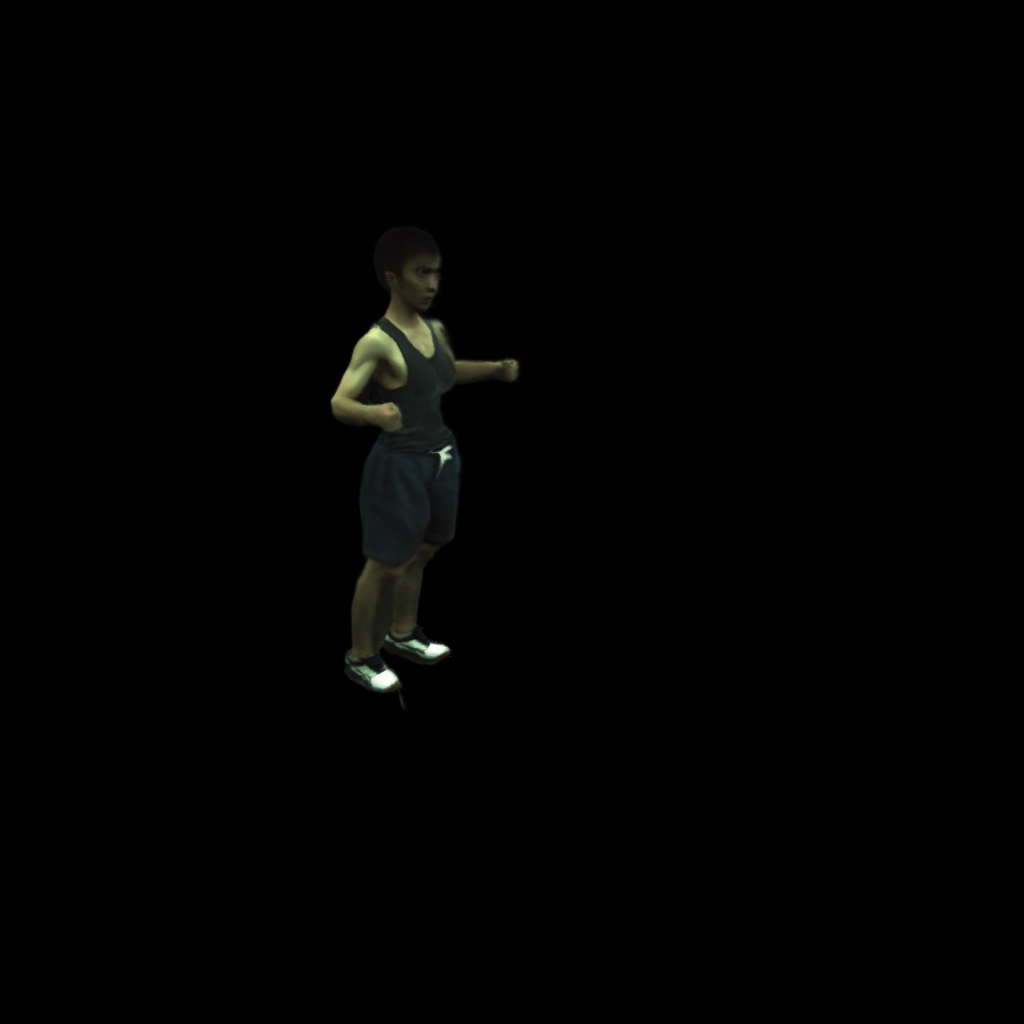}
    \end{subfigure}
    \begin{subfigure}[c]{0.0915\textwidth}
		\includegraphics[width=1\textwidth, trim=200 200 350 50, clip]{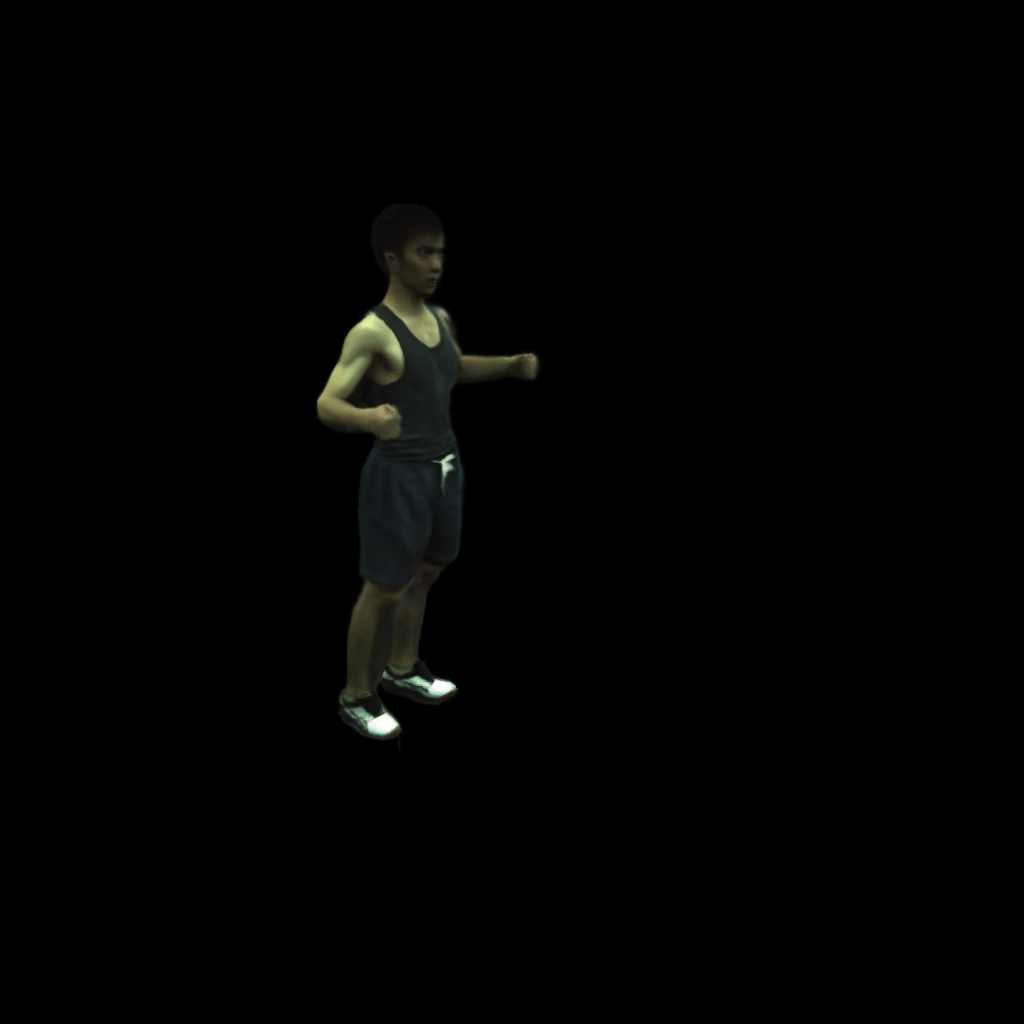}
    \end{subfigure}
    \begin{subfigure}[c]{0.0915\textwidth}
		\includegraphics[width=1\textwidth, trim=200 200 350 50, clip]{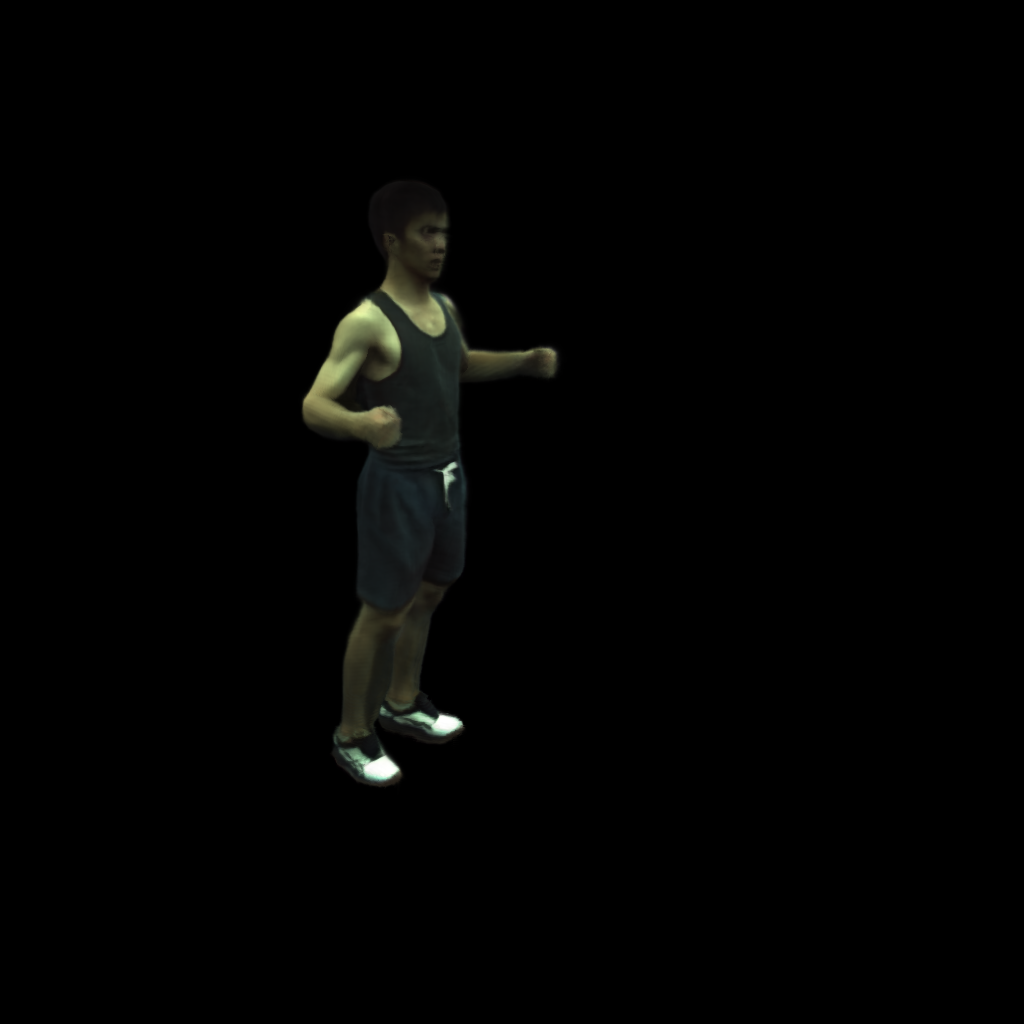}
    \end{subfigure}
    \begin{subfigure}[c]{0.0915\textwidth}
		\includegraphics[width=1\textwidth, trim=200 200 350 50, clip]{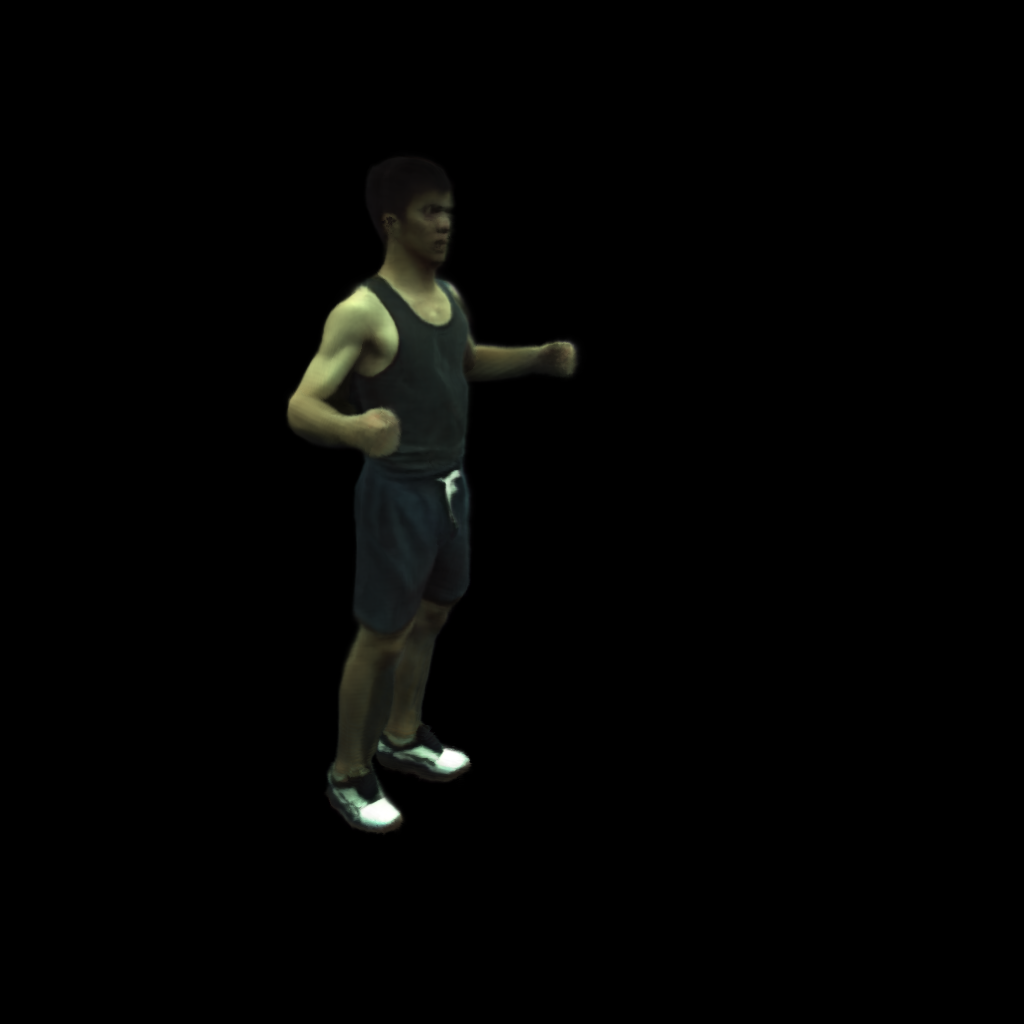}
    \end{subfigure}
    \begin{subfigure}[c]{0.0915\textwidth}
		\includegraphics[width=1\textwidth, trim=200 200 350 50, clip]{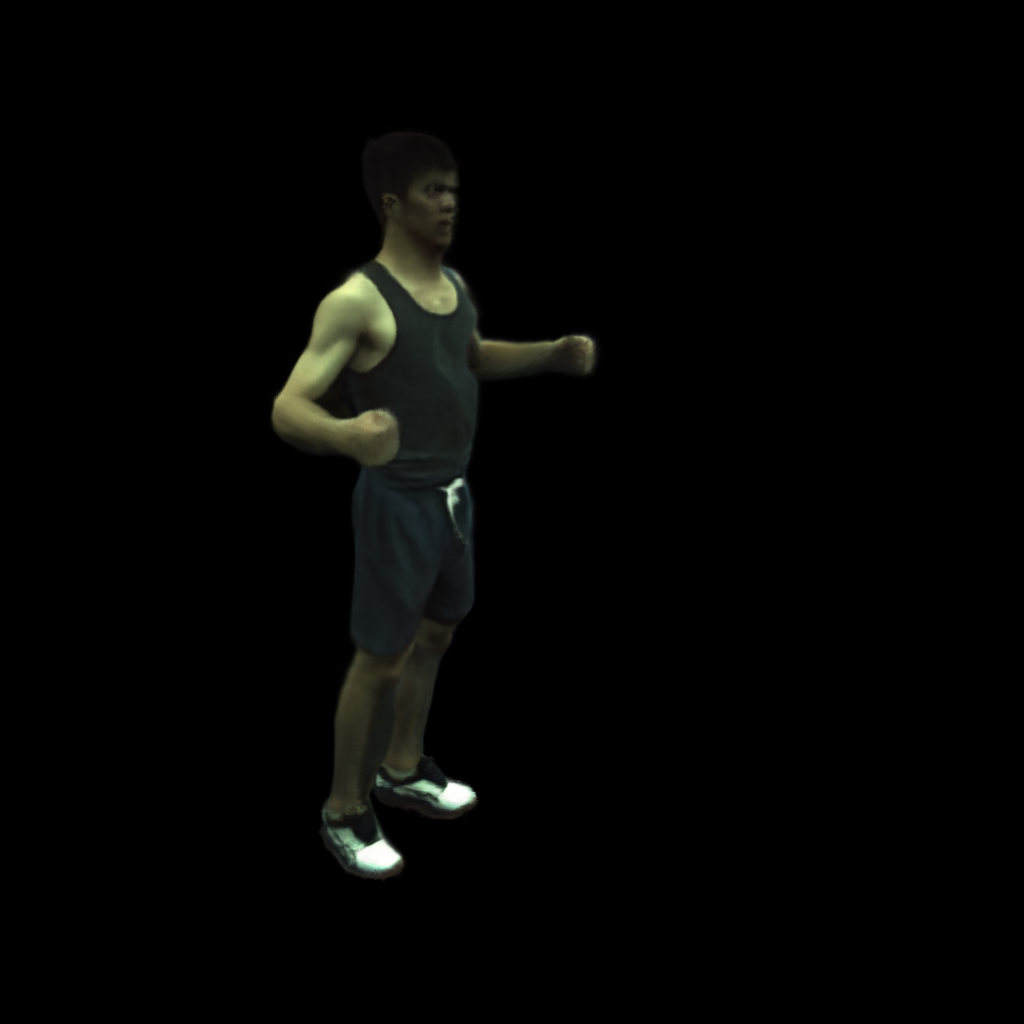}
    \end{subfigure}
  \vspace{-4pt}
 \caption{\textbf{Results of shape editing on subjects ``315'' and ``377'' of ZJU-MoCap.}}

\vspace{-2mm}
\label{fig:reshape}

\end{figure}

\vspace{0.5em}
\noindent \textbf{Metrics.} We follow previous work to evaluate our method on novel pose synthesis and relighting using three standard metrics, \ie, Peak Signal-to-Noise Ratio (PSNR), Structural Similarity Index Measure (SSIM) \cite{ssim}, and Learned Perceptual Image Patch Similarity (LPIPS) \cite{Zhang_2018_CVPR}. Akin to \cite{peng2021neural}, we employ a bounding box surrounding the human as mask for metric calculation. Note, for other edits allowed by our method (\eg, reshaping, retexturing, and reshadowing), we only provide visual results as there are no ground-truths.

\subsection{Comparison with State-of-the-art Methods}

\noindent \textbf{Evaluation on novel pose synthesis.}
We compare our method against the following five state-of-the-art methods on novel pose synthesis:
Neural Body (NB) \cite{peng2021neural}, Animatable NeRF (AN) \cite{peng2021animatable}, Dual-Space NeRF (DS) \cite{zhi2022dual}, ARAH \cite{ARAH:2022:ECCV}, and PoseVocab (PV) \cite{li2023posevocab}. For fair comparison, we produce their results using publicly-available implementation provided by the authors with recommended parameter setting. \cref{table:zjumocap} report the quantitative results, where we can see that our method outperforms the compared baselines in all three metrics. It is worth mentioning that our method also produces good novel view synthesis results. In addition, we show visual comparison in \cref{fig:com-zju}. Comparing the results, it is clear that our method can render photorealistic images with much more high-frequency details. 

\begin{figure}[!t]
    \centering
    \includegraphics[width=1.0\linewidth]{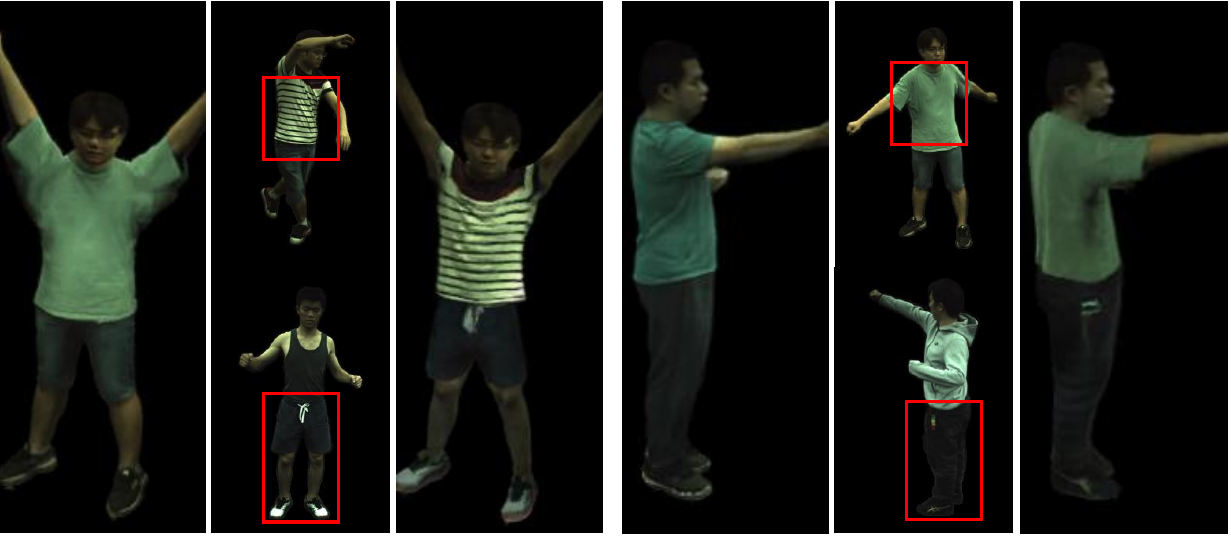}
    \vspace{-20pt}
    \caption{\textbf{Results of retexturing.} Our method allows to swap textures between body parts from different subjects.}
    \vspace{-2mm}
    \label{fig:recloth}
\end{figure}

\vspace{0.5em}
\noindent \textbf{Evaluation on relighting. } For relighting, we compare our method with Relighting4D \cite{chen2022relighting}, by naively placing the relighted human in a new background. We did not compare with \cite{zhen2023relightable,Sun_2023_ICCV}, since they have no publicly available code for now. \cref{fig:com-relight} gives the comparison results on relighting, as well as normal and albedo. As shown, our method generates physically realistic relighting results under novel pose, while Relighting4D fails to generate natural-looking results when the human in videos performs dramatic movement. \cref{fig:show-relight} shows more relighting results produced by our method, in both indoor and outdoor scenes. As can be seen, for all these cases, our method produces high-quality results.

\input{figs/reshadow}
\begin{figure}[tbp]
\centering
\captionsetup[subfigure]{labelformat=empty,labelsep=space}
    \begin{subfigure}[c]{0.115\textwidth}	
		\includegraphics[width=1\textwidth, trim=0 0 0 0, clip]{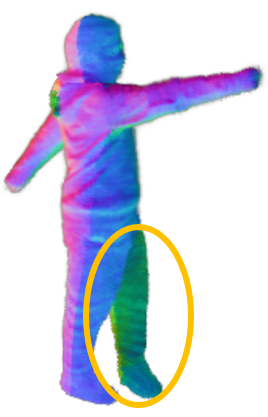}
  \vspace{-4mm}
    \caption{w/o normal reg.}
    \end{subfigure}
    \begin{subfigure}[c]{0.115\textwidth}	
		\includegraphics[width=1\textwidth, trim=0 0 0 0, clip]{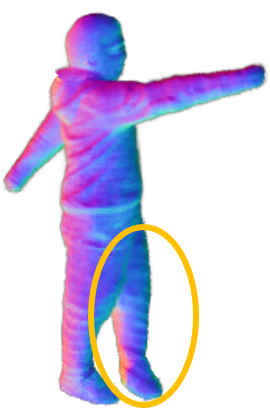}
  \vspace{-4mm}
    \caption{w/ normal reg.}
    \end{subfigure}
    \begin{subfigure}[c]{0.115\textwidth}	
		\includegraphics[width=1\textwidth, trim=0 0 0 0, clip]{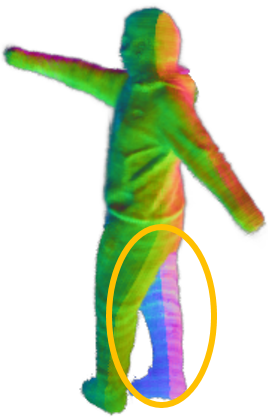}
  \vspace{-4mm}
    \caption{w/o normal reg.}
    \end{subfigure}
    \begin{subfigure}[c]{0.115\textwidth}	
		\includegraphics[width=1\textwidth, trim=0 0 0 0, clip]{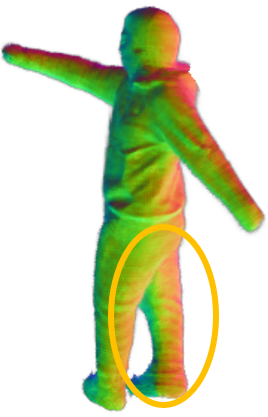}
  \vspace{-4mm}
    \caption{w/ normal reg.}
    \end{subfigure}
    \vspace{-2mm}
    \caption{\textbf{Effect of normal regularization $\mathcal{L}_n$.} As shown, adopting $\mathcal{L}_n$ benefits obtaining more accurate normal map.}
    \label{fig:ablation_normal}
\end{figure}

\begin{table}[tbp]
\centering
\resizebox{\linewidth}{!}{
\begin{tabular}{l c c  c  c}
\toprule[1pt]

 Model& Num of views & PSNR $\uparrow$ & SSIM $\uparrow$ & LPIPS $\downarrow$ \\ 
\midrule
w/o normal reg. & 4 & 26.6 & 0.928 & 0.084\\
w/o pose-aware & 4 & 26.6 & 0.929 & \textbf{0.078} \\
w/o subject-level & 4 & 26.6 & 0.928 & 0.079\\
full method & 1 & 24.2 & 0.906 & 0.117 \\
full method & 4 & \textbf{26.8} & \textbf{0.930} & 0.079\\
\bottomrule[1pt]

\end{tabular}
}
\vspace{-0.5em}
\caption{\textbf{Ablation studies on ZJU-MoCap.} ``w/o normal reg.'' refers to the exclusion of the proposed normal regularization $\mathcal{L}_n$, while ``w/o pose-aware'' indicates removal of the pose-aware feature, and ``w/o subject-level'' means that the subject-level feature is omitted. Note, here we report the average results of novel view and novel pose synthesis.}
\label{table:ablation-on-zjumocap}
 \vspace{-2mm}
\end{table}

\subsection{More Customization Results}
In addition to novel pose synthesis and relighting, our method also allows editing of shape, texture, and shadow. \cref{fig:reshape} shows our shape editing results, which are produced by adopting our local coordinate defined in tangent space to transform query points between different shape spaces. Following \cite{Xu_2022_CVPR}, we achieve retexturing by mapping the 3D query points from the target subject to the source subject. For each point in the canonical space of the target subject, we first adjust its position according to the bounding boxes of the two subjects, and then query the color from the canonical space of the source subject. By composing multiple learned neural human fields, we can achieve texture swapping, as shown in \cref{fig:recloth}. Moreover, \cref{fig:reshadow} shows that our method is applicable to shadow editing, and allows to eliminate self-cast shadows caused by clothing wrinkles and locally transfer shadows from one person to another. More customization results can be found in the supplementary material.

\input{figs/ablation}
\subsection{More Analysis}
\noindent\textbf{Ablation studies.}
Here we conduct ablation studies to validate the effectiveness of our subject-level feature, pose-aware feature, and normal regularization. Besides, we examine the performance of our method on single-view video. \cref{table:ablation-on-zjumocap} summarizes the quantitative results of ablation studies. \cref{fig:ablation_normal} presents visual comparison to verify the necessity of our normal regularization, while \cref{fig:ablation} demonstrates the benefit of subject-level and pose-aware features to our approach. Please see also the supplementary material for more results and analysis on different numbers of $R$ in Eq.~\eqref{tri-plane} and other possible alternatives (\ie, relative position, direction, and UVH) for our local tangent coordinate.

\vspace{0.5em}
\noindent\textbf{Limitations.} Since obtaining the subject-level feature requires to project the query point to the nearest point on SMPL faces, our performance can be sensitive to the estimated SMPL parameters. In addition, as we do not explicitly model visibility, shadows under complex novel poses may be erroneous. Moreover, we may generate blurred retexturing results due to large pose variance, and unrealistic reshaping results for extreme shape. Finally, similar to the vanilla NeRF, our method has to be optimized for each human separately.

\section{Conclusion}
\label{sec:discussion}
We have presented a novel framework called NECA for learning fully customizable human avatars from sparse-view or even monocular videos. In contrast to previous methods which offer limited editing capabilities, we provide neural avatar that allows high-fidelity editing on pose, viewpoint, lighting, shape, texture, and shadow. Extensive experiments validate the versatility and practicality of our approach, and our improvements over prior state-of-the-arts in novel pose synthesis and relighting. We hope that our work can shed light on the creation of customizable human avatar and its related applications.

\vspace{0.5em}
\noindent \textbf{Acknowledgement.} This work was supported by the National Natural Science Foundation of China (U21A20471, 62072191), Guangdong Basic and Applied Basic Research Foundation
(2023A1515030002, 2023B1515040025).

{\small
\bibliographystyle{ieeenat_fullname}
\bibliography{11_references}
}

\clearpage
\appendix \section{Implementation Details}
\label{sec:implementation}

\subsection{Building Tangent Space}
To construct the tangent space, we begin by computing the TBN (Tangent, Bitangent, Normal) matrix for the transformation. Consider a triangle with vertices $\mathbf{v}_1, \mathbf{v}_2, \mathbf{v}_3$, and edges $\mathbf{e}_1, \mathbf{e}_2, \mathbf{e}_3$. The normal of the triangle is computed as the cross product of edges $\mathbf{e}_1$ and $\mathbf{e}_2$:
\begin{equation}
    \mathbf{n}_{f} = \mathbf{e}_1 \times \mathbf{e}_2.
\end{equation}
Now, focusing on vertex $\mathbf{v}_1$ (similar computations apply to other vertices), the normal of vertex $\mathbf{v}_1$ can be calculated by:
\begin{equation}
    \mathbf{n}_1 = \frac{\sum_{m=1}^{N} \mathbf{n}_{f}^m}{\sum_{m=1}^{N} \|\mathbf{n}_{f}^m\|_2},
\end{equation}
where $N$ is the number of triangles that vertex $\mathbf{v}_1$ belongs to, and $\mathbf{n}_{f}^m$ is the normal vector of the $m$-th triangle. 
The tangent and bitangent vectors of $\mathbf{v}_1$ can be calculated by solving a combination of linear equations.
Given the texture coordinate $(u_i^*, v_i^*)$ of vertex $\mathbf{v}_i$, we express the edges as linear combinations:
\begin{equation}
    \begin{aligned}
        \mathbf{e}_1 &= \Delta u_1^* \mathbf{t}_1 + \Delta v_1^* \mathbf{b}_1, \\
        \mathbf{e}_2 &= \Delta u_2^* \mathbf{t}_1 + \Delta v_2^* \mathbf{b}_1.
    \end{aligned}
\end{equation}
This can also be written component-wise as:
\begin{equation}
    \begin{aligned}
        (e_{1x}, e_{1y}, e_{1z}) &= \Delta u_1^* (t_{1x}, t_{1y}, t_{1z}) + \Delta v_1^* (b_{1x}, b_{1y}, b_{1z}), \\
        (e_{2x}, e_{2y}, e_{2z}) &= \Delta u_2^* (t_{1x}, t_{1y}, t_{1z}) + \Delta v_2^* (b_{1x}, b_{1y}, b_{1z}),
    \end{aligned}
\end{equation}
where $\Delta u_1^*$ and $\Delta v_1^*$ are differences in texture coordinates along edge $\mathbf{e}_1$, $\Delta u_2^*$ and $\Delta v_2^*$ are differences along edge $\mathbf{e}_2$, and $\mathbf{t}_1$ and $\mathbf{b}_1$ represent the tangent and bitangent vectors for vertex $\mathbf{v}_1$.
These equations can be expressed as matrix multiplication:
\begin{equation}
    \begin{bmatrix} e_{1x} & e_{1y} & e_{1z} \\ e_{2x} & e_{2y} & e_{2z} \end{bmatrix} = \begin{bmatrix} \Delta u_1^* & \Delta v_1^* \\ \Delta u_2^* & \Delta v_2^* \end{bmatrix} \cdot \begin{bmatrix} t_{1x} & t_{1y} & t_{1z} \\ b_{1x} & b_{1y} & b_{1z} \end{bmatrix}.
\end{equation}
Solving for the tangent and bitangent vectors:
\begin{equation}
    \begin{bmatrix} t_{1x} & t_{1y} & t_{1z} \\ b_{1x} & b_{1y} & b_{1z} \end{bmatrix} = \begin{bmatrix} \Delta u_1^* & \Delta v_1^* \\ \Delta u_2^* & \Delta v_2^* \end{bmatrix}^{-1} \cdot \begin{bmatrix} e_{1x} & e_{1y} & e_{1z} \\ e_{2x} & e_{2y} & e_{2z} \end{bmatrix}.
\end{equation}

For an arbitrary surface point $\mathbf{x}_s$, its normal $\mathbf{n}_s$ can be computed by:

\begin{equation}
    \mathbf{n}_s = \mB_{u_s^*,v_s^*}(\mathbf{n}_1,\mathbf{n}_2,\mathbf{n}_3),
\end{equation}
where $\mB$ denotes barycentric interpolation, $(u_s^*,v_s^*)$ represents the UV coordinate of $\mathbf{x}_s$, and $\mathbf{n}_1, \mathbf{n}_2, \mathbf{n}_3$ are the normals of the three vertices of the triangle that $\mathbf{x}_s$ falls into. The computation of tangent and bitangent vectors is similar.

\subsection{Network Architecture}
\begin{figure}[th]
    \centering
    \includegraphics[width=\linewidth]{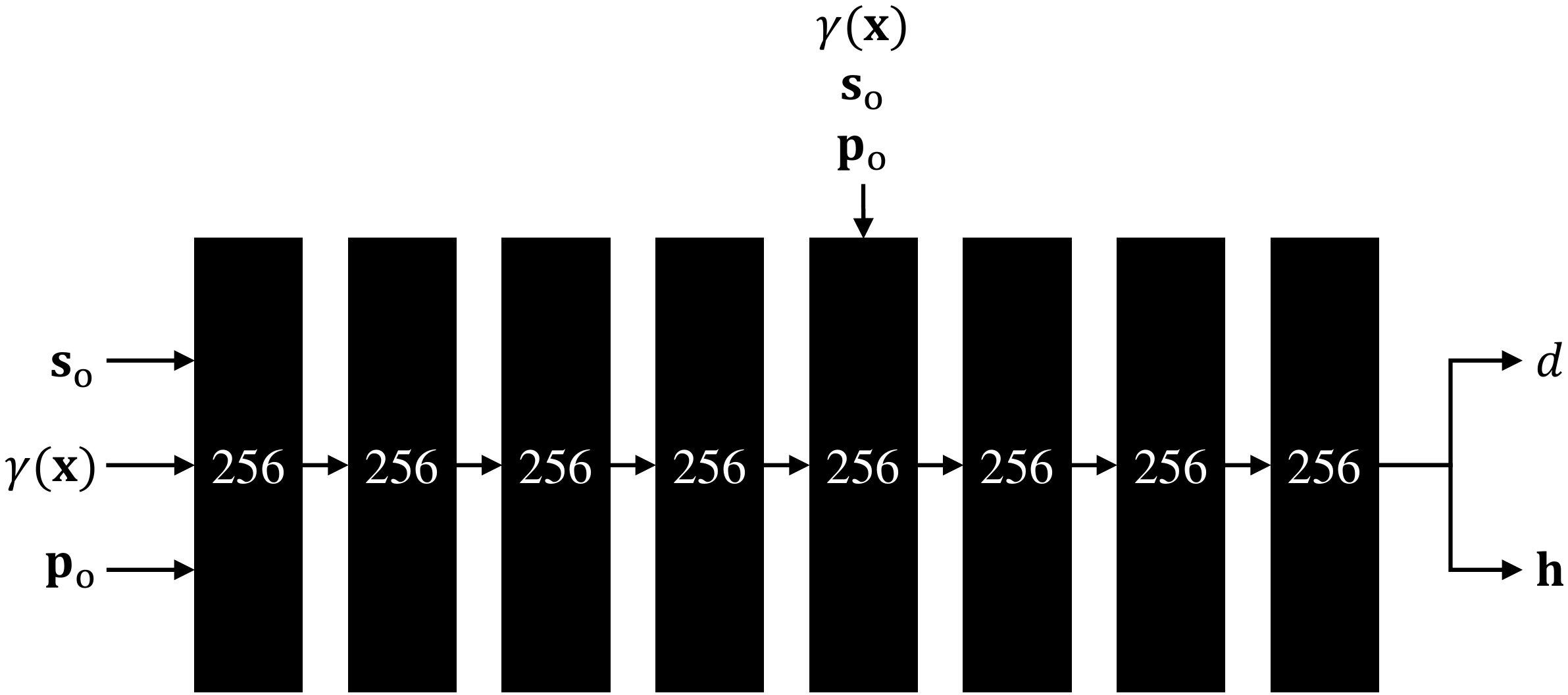}
    \caption{\textbf{SDF Network.} We show the architecture
    of our SDF network, which takes subject-level feature $\bs_o$, canonical position $\bx$, and pose-aware feature $\bp_o$ as input, and outputs signed-distance $d$ and latent feature vector $\bh$. A skip connection that concatenates $\gamma (\bx)$, $\bs_o$ and $\bp_o$ to the fifth layer is employed. Except for the last layer, each layer outputs 256 dimension features with softplus activations. }
    \label{fig:sdf_net}
\end{figure}

\begin{figure}[th]
    \centering
    \includegraphics[width=0.55\linewidth]{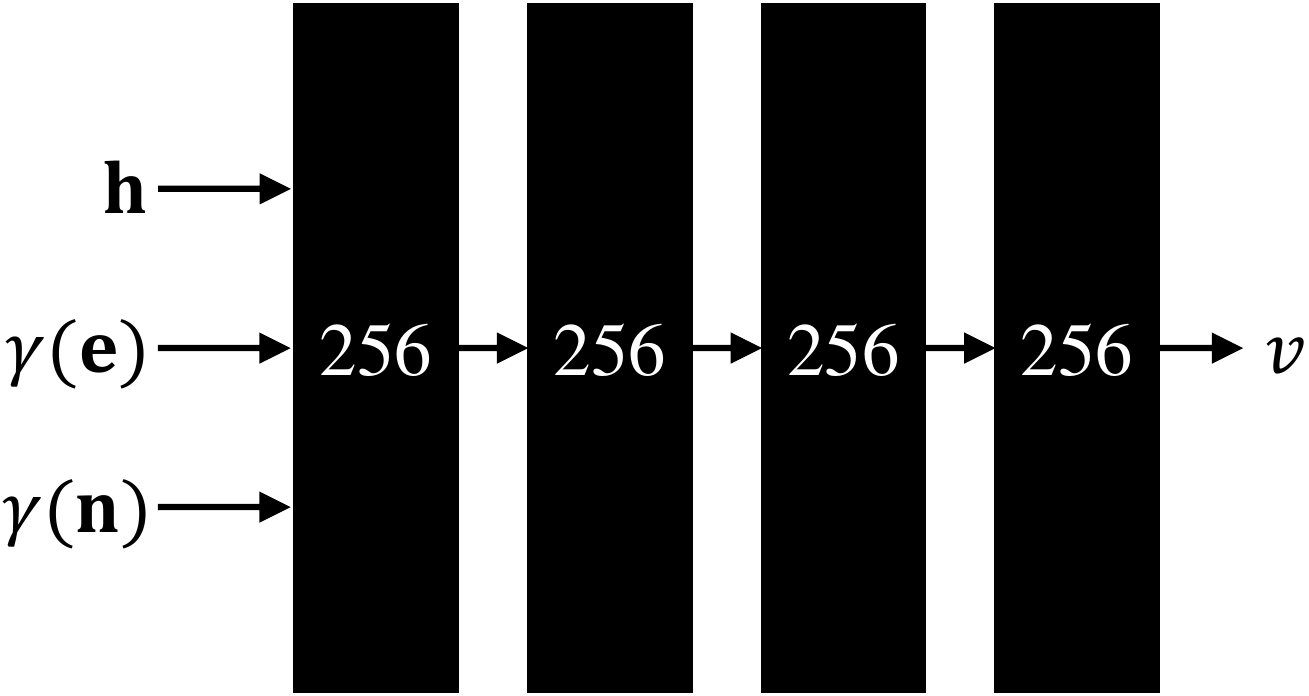}
    \caption{\textbf{Shadow Network.} We show the architecture
    of our shadow network, which takes latent feature $\bh$, viewing direction $\be$ and normal $\bn$ as input, and outputs shadow $v$.  Besides the last layer, each layer outputs 256 dimension features with ReLU activations. }
    \label{fig:shadow_net}
\end{figure}
\begin{figure}[th]
    \centering
    \includegraphics[width=0.55\linewidth]{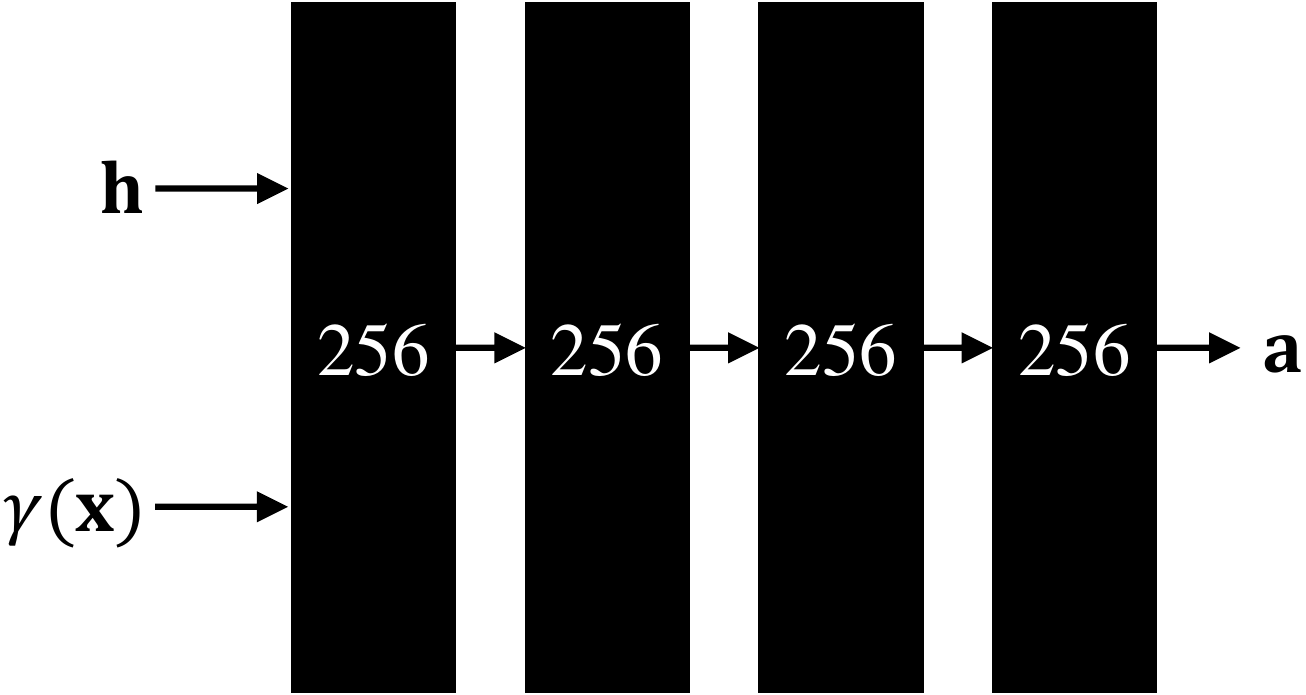}
    \caption{\textbf{Albedo Network.} We show the architecture
    of our albedo network, which takes latent feature $\bh$ and canonical position $\bx$ as input, and outputs albedo $\ba$. Each layer outputs 256 dimension features with ReLU activations, except for the last layer. }
    \label{fig:albedo_net}
\end{figure}
The detailed architecture of different sub-networks in our framework are illustrated in \cref{fig:sdf_net,fig:shadow_net,fig:albedo_net}. For canonical position $\bx$, viewing direction $\be$, and normal $\bn$, we employ positional encoding $\gamma$ to enhance the ability of capturing high-frequency details.

\begin{figure*}[th]
    \centering
    \includegraphics[width=\linewidth]{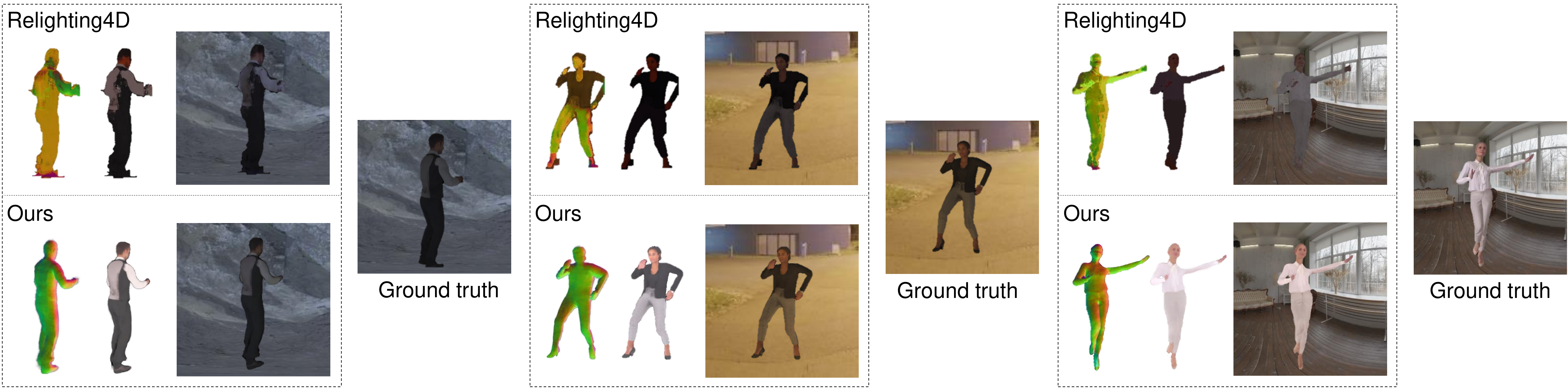}
    \vspace{-18pt}
    \caption{\textbf{Qualitative comparison of relighting on the synthetic dataset.} As can be seen, in contrast to the results of Relighting4D, our generated relighting results are visually closer to the ground truths. Best view in color.}
\label{fig:relight_on_syn}
\end{figure*}

\section{Relighting Results on Synthetic Dataset}
\subsection{Dataset Details}
To quantitatively evaluate the performance of our method on human relighting, we create a synthetic dataset with 12 videos using 3 publicly available 3D characters from \cite{renderpeople}, rigged with animations from \cite{Mixamo}. The dataset is rendered under 4 different lighting conditions, including one natural sunlight and three HDRI maps. Each monocular video has about 200 frames, where only 30 frames rendered under natural sunlight are used for training, the rest of frames are used for evaluation. Following previous work, we employ \cite{li2021hybrik} to estimate the SMPL and camera parameters.

\begin{table}[]
\centering
\resizebox{0.9\linewidth}{!}{
\begin{tabular}{l c  c  c}
\toprule[1pt]

 Model & PSNR $\uparrow$ & SSIM $\uparrow$ & LPIPS $\downarrow$ \\ 
\midrule
Relighting4D~\cite{chen2022relighting}& 21.9 & 0.811 & 0.201\\
Ours & \textbf{22.6} & \textbf{0.843} & \textbf{0.159} \\
\bottomrule[1pt]

\end{tabular}
}
\caption{\textbf{Quantitative comparison of relighting 
under novel pose on the synthetic dataset.} Our method outperforms Relighting4D on the synthetic dataset with ground truth relighting results. }
\label{table:compare_relighting}
\end{table}
\begin{table}[]
\centering
\resizebox{\linewidth}{!}{
\begin{tabular}{l  c  c  c  c  c c }
\toprule[1pt]
\multirow{2}{*}{Model} &  \multicolumn{3}{c}{Novel View} & \multicolumn{3}{c}{Novel Pose} \\ 
 & PSNR $\uparrow$ & SSIM $\uparrow$ & LPIPS $\downarrow$ & PSNR $\uparrow$ & SSIM $\uparrow$ & LPIPS $\downarrow $\\ 
\midrule
Rel. Pos. &28.0&0.955&0.052&26.0&0.936&0.067\\
Dir. &27.9&0.954&0.053&25.9&0.935&0.067\\
UVH &27.8&0.954&0.053&25.9&0.935&0.067\\
Ours &\textbf{28.3}&\textbf{0.957}&\textbf{0.051}&\textbf{26.3}&\textbf{0.940}&\textbf{0.066}\\
\bottomrule[1pt]
\end{tabular}}

\caption{\textbf{Ablation study on alternatives for our local tangent coordinate.} ``Rel. Pos.'' means relative position between sampled point and nearest point, while ``Dir.'' indicates direction and ``UVH'' represents barycentric coordinate and distance.}
\label{table:local_coords}
\end{table}
\begin{table}[]
\centering
\resizebox{\linewidth}{!}{
\begin{tabular}{l  c  c  c  c  c c c}
\toprule[1pt]
\multirow{2}{*}{Num of $R$} &  \multicolumn{3}{c}{Novel View} & \multicolumn{3}{c}{Novel Pose}&\multirow{2}{*}{Param.}  \\ 
 & PSNR $\uparrow$ & SSIM $\uparrow$ & LPIPS $\downarrow$ & PSNR $\uparrow$ & SSIM $\uparrow$ & LPIPS $\downarrow $&\\ 
\midrule
1 &27.9&0.953&0.051&25.8&0.934&0.066&1.70M\\
12 &28.0&0.954&0.051&25.9&0.935&0.066&5.82M\\
48 &28.3&\textbf{0.957}&0.051&\textbf{26.3}&\textbf{0.940}&0.066&19.30M\\
64 &\textbf{28.4}&\textbf{0.957}&0.051&\textbf{26.3}&\textbf{0.940}&0.066&25.30M\\
\bottomrule[1pt]
\end{tabular}}

\caption{\textbf{Ablation study on the number of $R$.} As shown, larger $R$ corresponds to overall better performance on novel view and pose synthesis, but this trend becomes less obvious when $R \geq 48$. To trade off the performance and the model size, $R=48$ is utilized in all our experiments.}
\label{table:num_R}
\end{table}

\subsection{Quantitative and Qualitative Results}
\cref{table:compare_relighting} demonstrates that our method clearly outperforms Relighting4D~\cite{chen2022relighting} on the synthetic dataset. Visual comparison results are presented in \cref{fig:relight_on_syn}, where our method exhibits superior performance on relighting, even in dealing with a challenging dataset with only 30 frames for training. In comparison, Relighting4D~\cite{chen2022relighting} struggles to produce physically convincing relighting renderings.

\section{Additional Quantitative Results}
\begin{table}[]
\centering

\resizebox{\linewidth}{!}{
\begin{tabular}{l c  c  c  c  c c }
\toprule[1pt]
\multirow{2}{*}{} & \multicolumn{3}{c}{Novel View} & \multicolumn{3}{c}{Novel Pose}\\ 
 & PSNR $\uparrow$ & SSIM $\uparrow$ & LPIPS $\downarrow$ & PSNR $\uparrow$ & SSIM $\uparrow$ & LPIPS $\downarrow $ \\ 
\midrule
NB~\cite{peng2021neural} &19.3&0.889&0.129&17.4&0.863&0.151\\
AN~\cite{peng2021animatable} &17.8&0.875&0.154&16.7&0.855&0.164\\
ARAH~\cite{ARAH:2022:ECCV} &19.5&0.893&0.124&17.8&0.868&0.144\\
PV~\cite{li2023posevocab} &20.2&0.903&0.105&18.0&0.870&0.121\\
Ours &\textbf{20.9}&\textbf{0.910}&\textbf{0.100}&\textbf{19.1}&\textbf{0.883}&\textbf{0.117}\\ 

\bottomrule[1pt]
\end{tabular}
}
 \vspace{-3mm}
\caption{\textbf{More quantitative comparison on DeepCap and DynaCap datasets.}}
\label{table:more_quan}
\vspace{-3mm}
\end{table}
\begin{table}[]
\centering

\resizebox{\linewidth}{!}{
\begin{tabular}{l c  c  c  c  c c }
\toprule[1pt]
\multirow{2}{*}{} & \multicolumn{3}{c}{Novel View} & \multicolumn{3}{c}{Novel Pose}\\ 
 & PSNR $\uparrow$ & SSIM $\uparrow$ & LPIPS $\downarrow$ & PSNR $\uparrow$ & SSIM $\uparrow$ & LPIPS $\downarrow $ \\ 
\midrule
HN~\cite{weng_humannerf_2022_cvpr} &30.4&0.974&0.024&23.8&0.936&0.069\\  
Ours& \textbf{30.9}&\textbf{0.978}&\textbf{0.023}&\textbf{30.9}&\textbf{0.977}&\textbf{0.023}\\

\bottomrule[1pt]
\end{tabular}
}
 \vspace{-3mm}
\caption{\textbf{Quantitative comparison with HumanNeRF on subject ``377'' in ZJU-Mocap.}}
\label{table:mono}
\vspace{-3mm}
\end{table}

We present additional quantitative comparison results on the DeepCap~\cite{deepcap} and DynaCap~\cite{habermann2021dynacap} datasets in \cref{table:more_quan}. Specifically, we evaluate the performance of "lan" from DeepCap and "vlad" from DynaCap. For each subject, we employ 300 frames for training and another 300 frames for evaluation. Notably, our method demonstrates superior performance over all baseline models by a significant margin.

Additionally, we conduct a comparison with the monocular method HumanNeRF~\cite{weng_humannerf_2022_cvpr} in \cref{table:mono}. Our results indicate that our approach achieves superior performance, particularly in the context of novel pose synthesis.
\section{Additional Ablation Studies}
\begin{figure*}[ht]  
  \centering    

   \includegraphics[width=1\textwidth]{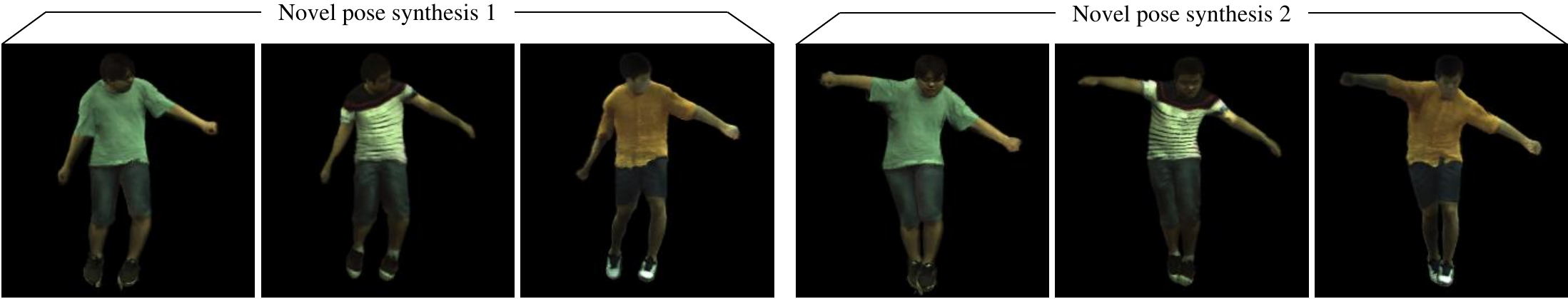}
 \caption{\textbf{More reposing results on ZJU-MoCap dataset.} We show two groups of novel pose synthesis results with poses from AIST++ \cite{li2021learn}. }
\label{fig:animation}
\end{figure*}
\input{figs/local_coords}
\subsection{Ablation Study on Local Coordinate}
As described in our paper, we introduce a novel local coordinate defined in tangent space. To validate the effectiveness of our local coordinate, we compare it with three alternatives. One common choice is the relative position. Denoting the sampled point as $\bx_o$ and the nearest point on SMPL as $\bx_s^*$, the relative position $\bc_{pos}$ is expressed as $\bc_{pos} = \bx_o - \bx_s^*$. Another alternative is the direction $\bc_{dir}$, defined as the normalization of the relative position: $\bc_{dir} = \frac{\bc_{pos}}{\|\bc_{pos}\|_2}$. The third option is UVH, which includes barycentric coordinate and distance: $\bc_{uvh} = (u,v,h)$, where $(u,v)$ denotes the barycentric coordinate, $h$ represents the distance between $\bx_o$ and $\bx_s^*$. To examine the performance of these alternatives, we conduct evaluation on the ``377'' subject in ZJU-MoCap~\cite{peng2021neural}, and provide the quantitative results in \cref{table:local_coords}. Comparing the results, we can see that our proposed local coordinate achieves the best results on all three metrics, manifesting its effectiveness and superiority.  In addition, we present the qualitative comparison results in \cref{fig:local_coord}, where our local coordinate in tangent space exhibits better alignment effect.

\input{figs/num_R}
\subsection{Ablation Study on Component Number $R$}
To assess the influence of different component numbers $R$ in the CP decomposition, we conduct experiments on the subject ``377'' from ZJU-MoCap. The numerical results for novel view and pose synthesis are summarized in \cref{table:num_R}. As shown, higher component number $R$ corresponds to better performance, but incurs larger model size. We opt for $R=48$ in our method, for striking a balance between the performance and the model size. The visual comparison results on different $R$ are shown in \cref{fig:num_R}, where we can see that the generated renderings with larger $R$ are visually better with less visual artifacts. Note that, the renderings with $R=48$ and $R=64$ are visually similar, showing that $R=64$ would not induce noticeable performance increase compared to $R=48$. 

\section{More Customization Results}
\begin{figure*}[th]
    \centering
    \includegraphics[width=\linewidth]{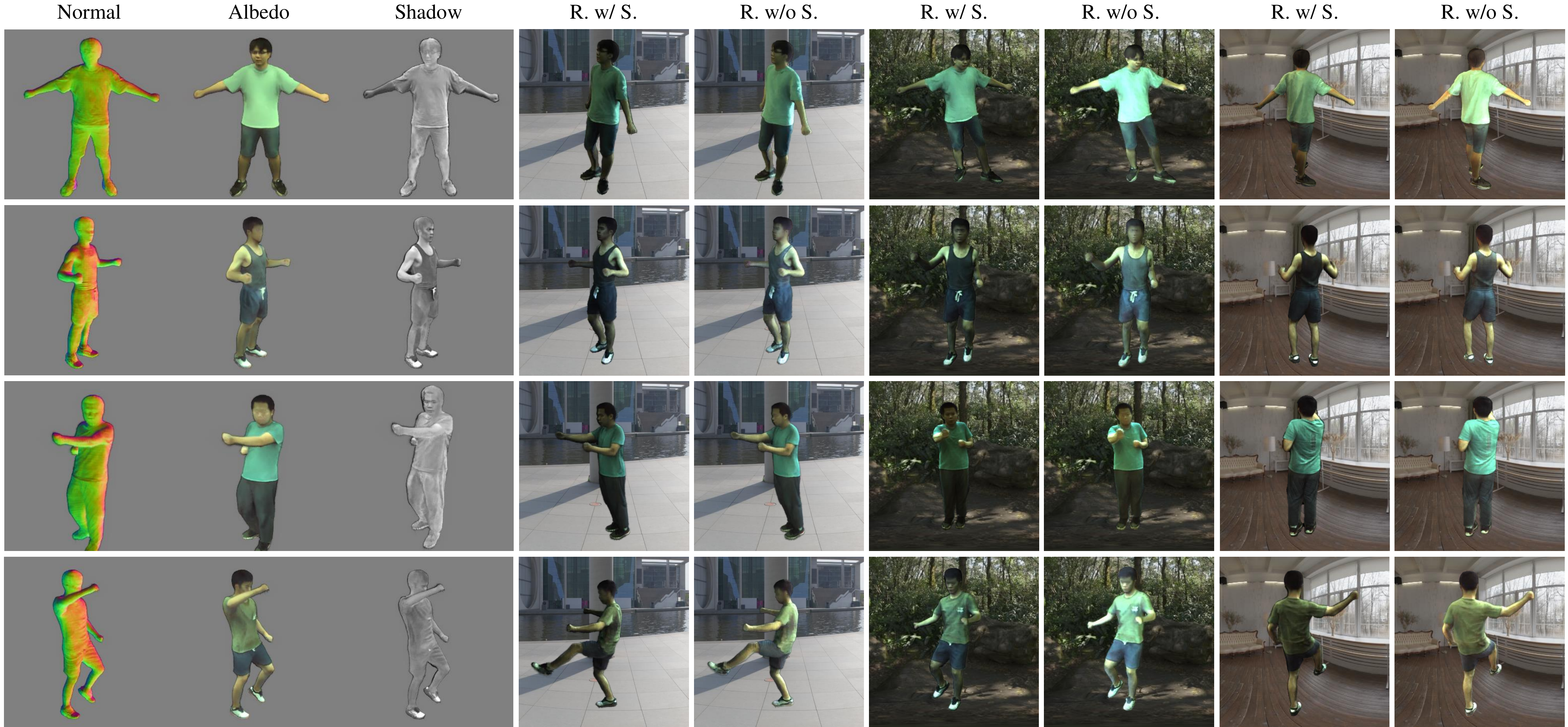}
\caption{\textbf{More relighting results on ZJU-MoCap dataset.} ``R. w/ S.'' and ``R. w/o S.'' refer to results with and without shadows.}
\label{fig:more_relight}
\end{figure*}

\begin{figure*}[th]
    \centering
    \includegraphics[width=\linewidth]{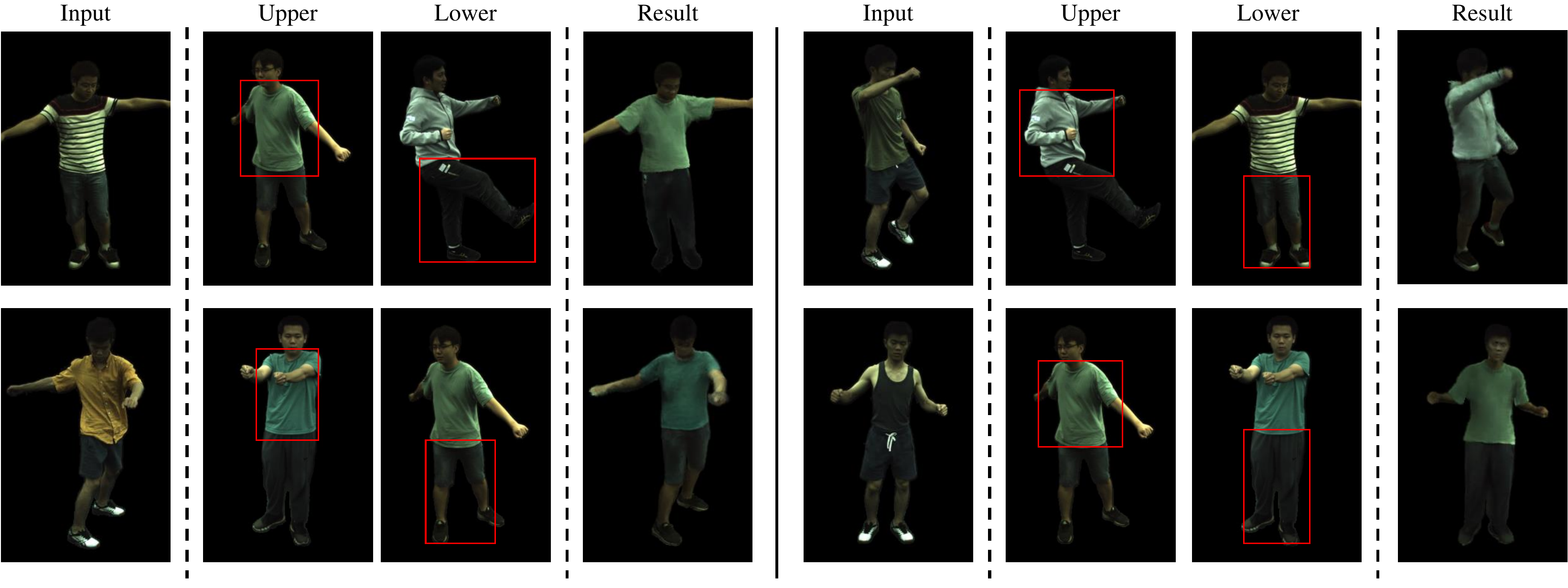}
    \vspace{-3mm}
\caption{\textbf{More retexturing results on ZJU-MoCap dataset.}}
\vspace{-1mm}
\label{fig:more_recloth}
\end{figure*}

\begin{figure*}[ht]  
  \centering    
  \captionsetup[subfigure]{labelformat=empty,labelsep=space}
    \begin{subfigure}[c]{0.16\textwidth}		\includegraphics[width=1\textwidth, trim=100 100 200 50, clip]{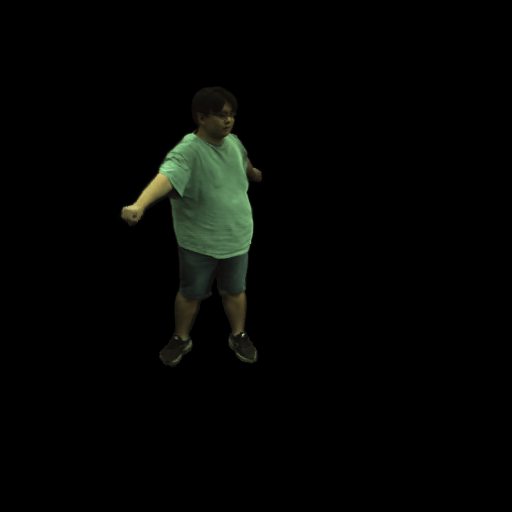}
    \end{subfigure}
    \begin{subfigure}[c]{0.16\textwidth}		\includegraphics[width=1\textwidth, trim=100 100 200 50, clip]{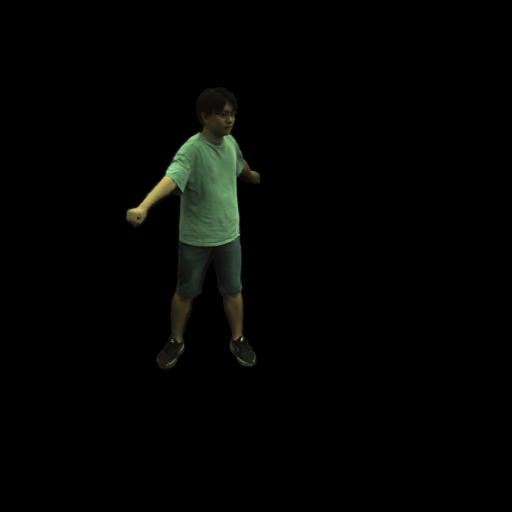}
    \end{subfigure}
    \begin{subfigure}[c]{0.16\textwidth}		\includegraphics[width=1\textwidth, trim=100 100 200 50, clip]{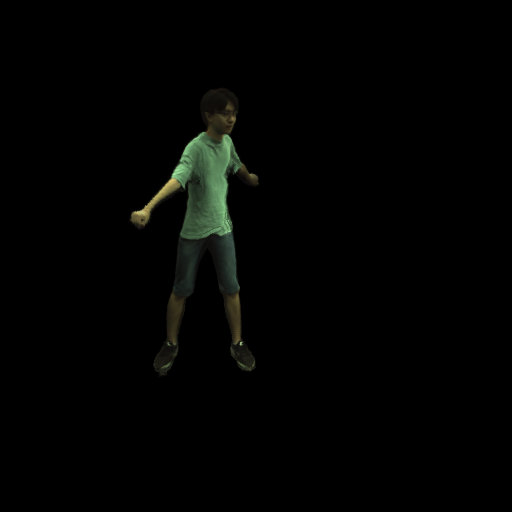}
    \end{subfigure}
    \begin{subfigure}[c]{0.16\textwidth}		\includegraphics[width=1\textwidth, trim=100 80 200 70, clip]{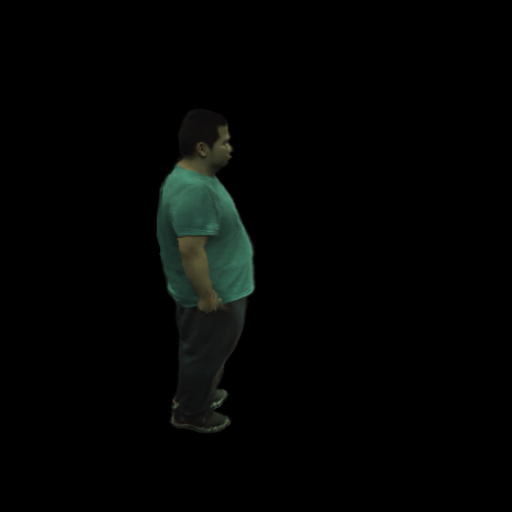}
    \end{subfigure}
    \begin{subfigure}[c]{0.16\textwidth}		\includegraphics[width=1\textwidth, trim=100 80 200 70, clip]{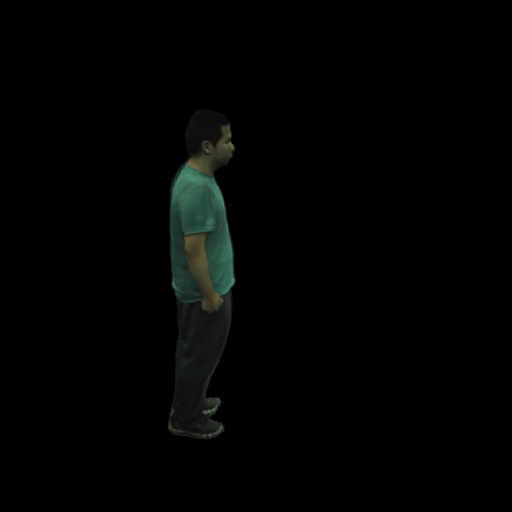}
    \end{subfigure}
    \begin{subfigure}[c]{0.16\textwidth}		\includegraphics[width=1\textwidth, trim=100 80 200 70, clip]{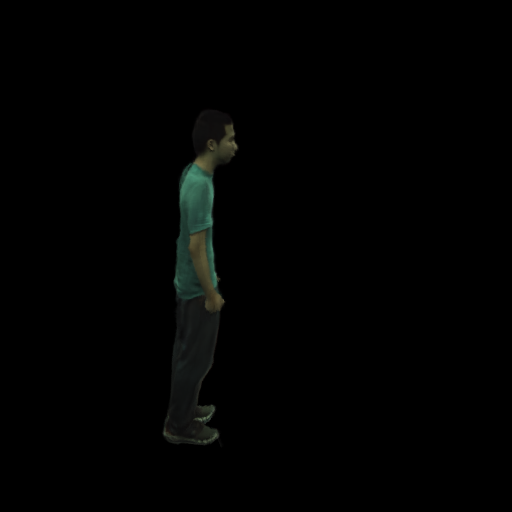}
    \end{subfigure}\vspace{1pt}\\
    \begin{subfigure}[c]{0.16\textwidth}		\includegraphics[width=1\textwidth, trim=100 100 200 50, clip]{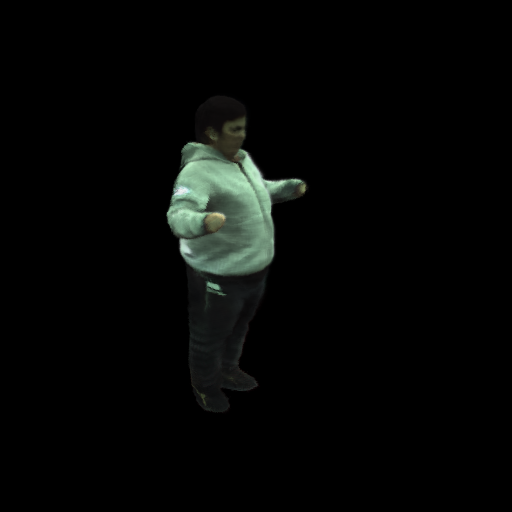}
    \end{subfigure}
    \begin{subfigure}[c]{0.16\textwidth}		\includegraphics[width=1\textwidth, trim=100 100 200 50, clip]{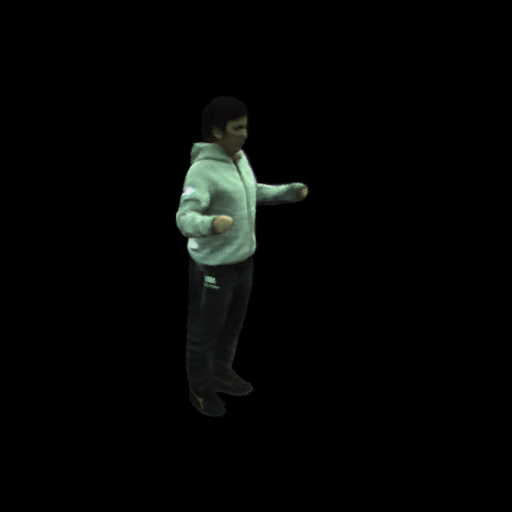}
    \end{subfigure}
    \begin{subfigure}[c]{0.16\textwidth}		\includegraphics[width=1\textwidth, trim=100 100 200 50, clip]{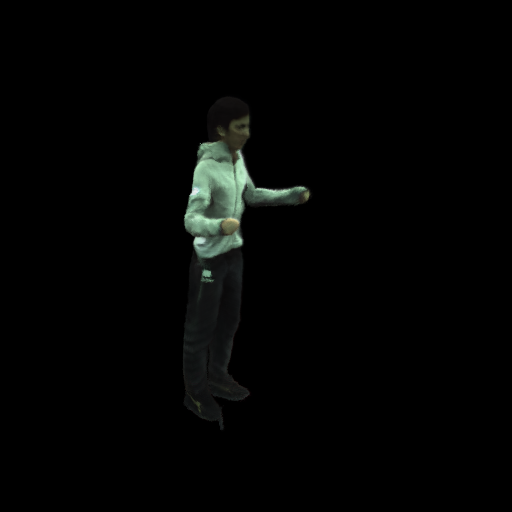}
    \end{subfigure}
    \begin{subfigure}[c]{0.16\textwidth}		\includegraphics[width=1\textwidth, trim=100 80 200 70, clip]{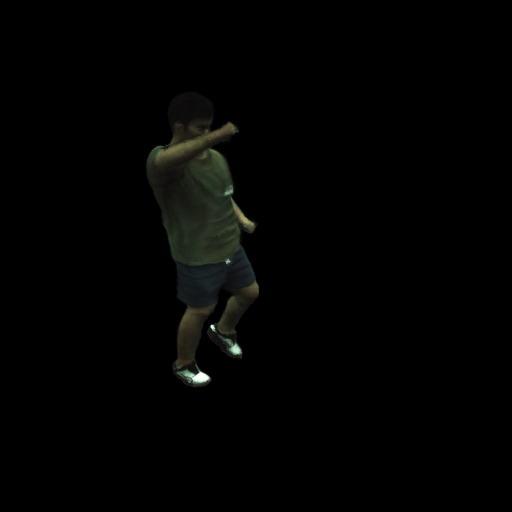}
    \end{subfigure}
    \begin{subfigure}[c]{0.16\textwidth}		\includegraphics[width=1\textwidth, trim=100 80 200 70, clip]{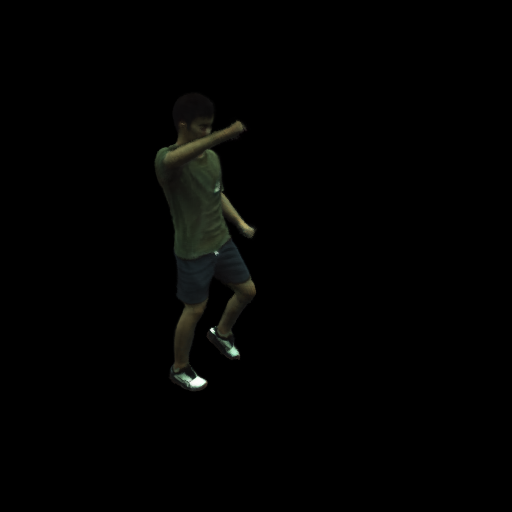}
    \end{subfigure}
    \begin{subfigure}[c]{0.16\textwidth}		\includegraphics[width=1\textwidth, trim=100 80 200 70, clip]{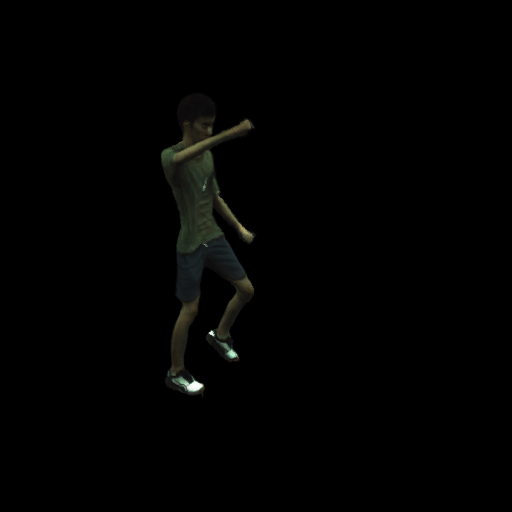}
    \end{subfigure}\vspace{1pt}\\
    \begin{subfigure}[c]{0.16\textwidth}		\includegraphics[width=1\textwidth, trim=100 100 200 50, clip]{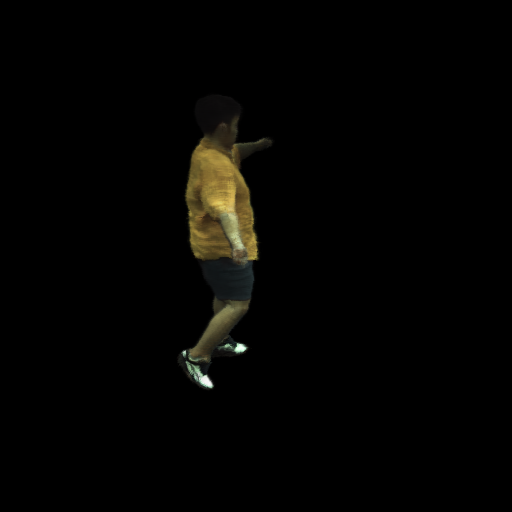}
    \end{subfigure}
    \begin{subfigure}[c]{0.16\textwidth}		\includegraphics[width=1\textwidth, trim=100 100 200 50, clip]{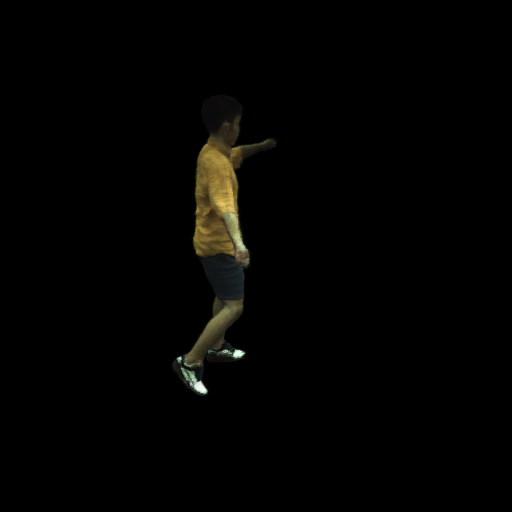}
    \end{subfigure}
    \begin{subfigure}[c]{0.16\textwidth}		\includegraphics[width=1\textwidth, trim=100 100 200 50, clip]{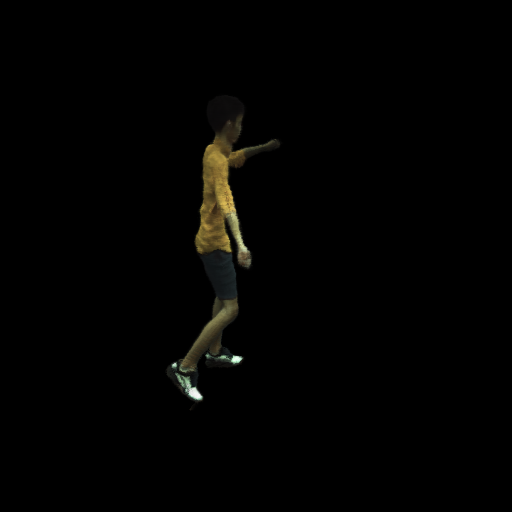}
    \end{subfigure}
    \begin{subfigure}[c]{0.16\textwidth}		\includegraphics[width=1\textwidth, trim=100 80 200 70, clip]{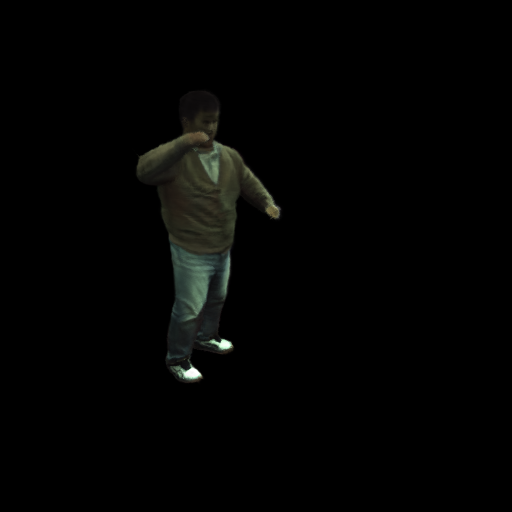}
    \end{subfigure}
    \begin{subfigure}[c]{0.16\textwidth}		\includegraphics[width=1\textwidth, trim=100 80 200 70, clip]{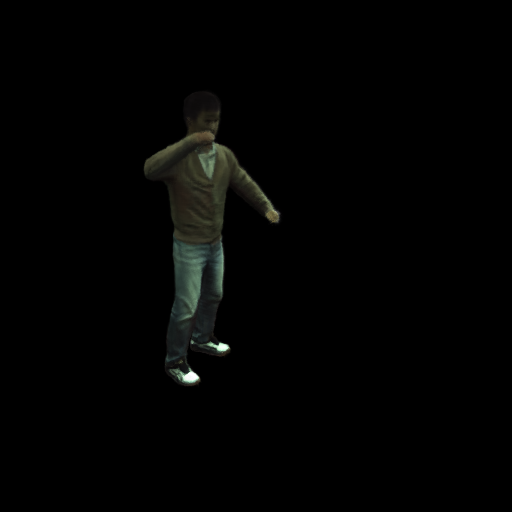}
    \end{subfigure}
    \begin{subfigure}[c]{0.16\textwidth}		\includegraphics[width=1\textwidth, trim=100 80 200 70, clip]{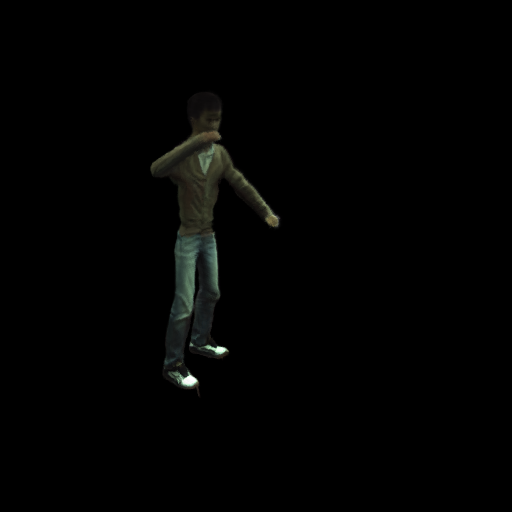}
    \end{subfigure}
    
 \caption{\textbf{More shape editing results on ZJU-MoCap dataset.}}
\label{fig:more_reshape}
\end{figure*}

We also showcase additional results in \cref{fig:animation,fig:more_relight,fig:more_reshape,fig:more_recloth,fig:more_reshadow} to further highlight the versatility of our method in various customization tasks. For a comprehensive overview, we encourage reviewers to refer to the supplementary video.

\input{figs/more_reshadow}

\section{Ethics Statement}

The datasets utilized in our research are sourced from publicly available repositories, including ZJU-MoCap \cite{peng2021neural}, NeuMan \cite{jiang2022neuman}, DynaCap \cite{habermann2021dynacap} and DeepCap \cite{deepcap}. Additionally, we incorporate a dataset generated using 3D characters from RenderPeople \cite{renderpeople}. The meticulous collection of these datasets aligns with ethical guidelines and principles.

Given the capabilities of our work in generating realistic animated humans, it is crucial to acknowledge potential ethical considerations. NECA has the capacity to depict humans in various scenarios, including wearing different clothing with realistic lighting and shadow effects. While this technology offers creative possibilities, it also raises concerns about the potential misuse of generating misleading or fabricated videos of real individuals, contributing to negative social impacts. To address this, we emphasize the responsible and ethical use of our technology. Moreover, we recognize the environmental impact of the computational resources required by our method, which could contribute to increased carbon emissions. In response, we commit to releasing our pretrained weights to promote computational efficiency and reduce the environmental footprint associated with using our approach. This proactive measure aims to balance technological advancements with ethical considerations.

\end{document}